\documentclass[12pt]{article}

\usepackage{scicite}

\usepackage{times}
\usepackage{tabularx}

\usepackage{dsfont}
\usepackage{amsmath}
\usepackage{amssymb}
\usepackage{multirow}
\usepackage{mathtools}
\usepackage{units}
\usepackage{subfloat}
\usepackage{float}
\usepackage[]{units}
\usepackage{xcolor}
\usepackage{url}
\usepackage{bbm}
\usepackage{algorithm}
\usepackage{algpseudocode}
\usepackage{soul} 
\usepackage[makeroom]{cancel}
\usepackage[printonlyused]{acronym}
\usepackage{hyperref}
\usepackage{makecell}
\usepackage{lineno}
\usepackage{array}
\usepackage{ragged2e}
\usepackage{threeparttable}
\usepackage[most]{tcolorbox}

\usepackage{caption}
\usepackage[printonlyused]{acronym}
\newacro{LMHV}{Low-Mix High-Volume}
\newacro{HMLV}{High-Mix Low-Volume}
\newacro{DOF}{Degree Of Freedom}
\newacro{BC}{Behavior Cloning}
\newacro{EE}{End Effector}
\newacro{IL}{Imitation Learning}
\newacro{RL}{Reinforcement Learning}
\newacro{VLA}{Vision Language Action model}
\newacro{ACT}{Action Chunking with Transformers}
\newacro{VFM}{Vision Foundation Model}
\newacro{SSM}{Speed and Separation Monitoring}
\newacro{FSM}{Finite State Machine}
\newacro{PFL}{Power Force Limiting}
\newacro{QC}{Quality Control}
\newacro{SAM2}{Segment Anything Model 2}

\newtcolorbox{promptbox}{
    colback=gray!5,
    colframe=black!40,
    boxrule=0.4pt,
    arc=1mm,
    left=2mm,
    right=2mm,
    top=1mm,
    bottom=1mm
}

\newenvironment{sciabstract}{%
\begin{quote} \bf}
{\end{quote}}

\title{Learning-augmented robotic automation \\for real-world manufacturing}

\author{
Yunho Kim\footnote{Corresponding authors: yunho.kim@neuromeka.com, joonho.lee@neuromeka.com}, Van Quan Nguyen, Taewhan Kim, Young Jin Heo, Joonho Lee* \\
\normalsize{All authors are with Neuromeka Co., Ltd., Seoul, Republic of Korea}
}

\date{}

\begin{document} 

\baselineskip24pt

\maketitle 

\begin{sciabstract}
Industrial robots are widely used in manufacturing, yet most manipulation still depends on fixed waypoint scripts that are brittle to environmental changes. Learning-based control offers a more adaptive alternative, but it remains unclear whether such methods, still mostly confined to laboratory demonstrations, can sustain hours of reliable operation, deliver consistent quality, and behave safely around people on a live production line. Here we present Learning-Augmented Robotic Automation, a hybrid system that integrates learned task controllers and a neural 3D safety monitor into conventional industrial workflows. It retains the core industrial automation backbone while introducing learning only where adaptability is needed. We deployed the system on an electric-motor production line to automate deformable cable insertion and soldering under real manufacturing constraints, a step previously performed manually by human workers. With less than 20 minutes of real-world data per task, the system operated continuously for 5 hours 10 minutes, producing 108 motors without physical fencing and achieving a 99.4\% pass rate on product-level quality-control tests. It achieved near-human takt time with reduced cycle-time variability while consistently producing high-quality solder joints comparable to those of experienced operators. These results establish a practical pathway for extending industrial automation with learning-based methods.
\end{sciabstract}

Industrial robots are widely used in manufacturing~\cite{vysocky2016cobotMinOverview, el2019cobotProgramReview, graetz2018robots}. However, they are mostly deployed through waypoint-based “teaching” procedures, in which engineers specify fixed sequences of pose targets and discrete actions (e.g., grasping or tool activation) using standardized programming interfaces~\cite{urManual, nrmkManual}. This paradigm enables reliable execution in structured settings, but it is inherently limited when tasks involve part-to-part variation or environmental uncertainty, because the robot largely replays pre-recorded motions with little perceptual feedback~\cite{cong2021visionReview}. As a result, several production steps remain manual not because they are wholly unstructured, but because they involve geometric variation, deformable materials, or tight tolerance.

Recent advances in learning-based control offer a promising alternative by enabling robots to acquire perception-driven, closed-loop manipulation skills directly from data. Such approaches allow robots to perform complex, contact-rich manipulation while reducing reliance on manually engineered perception, planning, and control pipelines. Neural controllers obtained with imitation learning~\cite{zhao2023ACT, fu2024mobileACT, chi2024diffusionpolicy, kim2025srtH}, reinforcement learning~\cite{luo2024serl, luo2024hilserl, lei2025rl100}, or large-scale multimodal training~\cite{rt22023rt2, kim24openvla, black2024pi0, intelligence2025pi0.5, bjorck2025grootN1, team2025geminiRobotics} have demonstrated success on complex manipulation tasks that are difficult or impractical to solve using conventional waypoint-based programming. However, despite these advances, their applicability to real-world industrial systems remains unclear.

Industrial deployment of learning-based controllers imposes stringent requirements. First, controllers must exhibit long-horizon stability with near-perfect success rates under strict \ac{QC} standards. Second, cycle time must be comparable to human takt time. Third, controllers must ensure safety in shared workspaces by reacting to human activity, thereby enabling an efficient human-robot division of labor. Finally, these requirements must be achieved under a limited training data budget and safety constraints, as large-scale data collection and risky exploratory motions~\cite{li2026failure} impose significant operational burdens on field engineers.

Existing learning-based methods often fall short of these requirements. They often exhibit limited success rates, slow execution, and substantial task-specific data requirements (often ranging from several to hundreds of hours)~\cite{ma2024surveyVLA, tsuji2025surveyIL, dreczkowski2025learning, ranawaka2025sail}. Furthermore, the deployment of such neural controllers in industrial automation settings remains underexplored, especially for full workstation-level operation. As a result, critical considerations—including safe human–robot coordination on the production line and consistent satisfaction of \ac{QC} standards during long-horizon operation—are rarely addressed, as most prior academic demonstrations remain limited to controlled laboratory environments and isolated tasks.

In this work, we present Learning-Augmented Robotic Automation, a factory-validated hybrid system that integrates reliable conventional automation with learning-based control in a safety-aware architecture (\href{https://youtu.be/8Uu_UGyLyKY}{Supplementary Movie 1}). Instead of relying on a single end-to-end policy, our approach retains the core industrial backbone—an explicit task scheduler and pre-taught motions for structured parts of the workflow— while introducing learning only where adaptability is required. This design preserves the predictability, precision, and robustness of classical control in structured sub-tasks while enabling perception-driven adaptation where it is most needed.

\begin{figure}[t!]
\centering
\includegraphics[width=\textwidth]{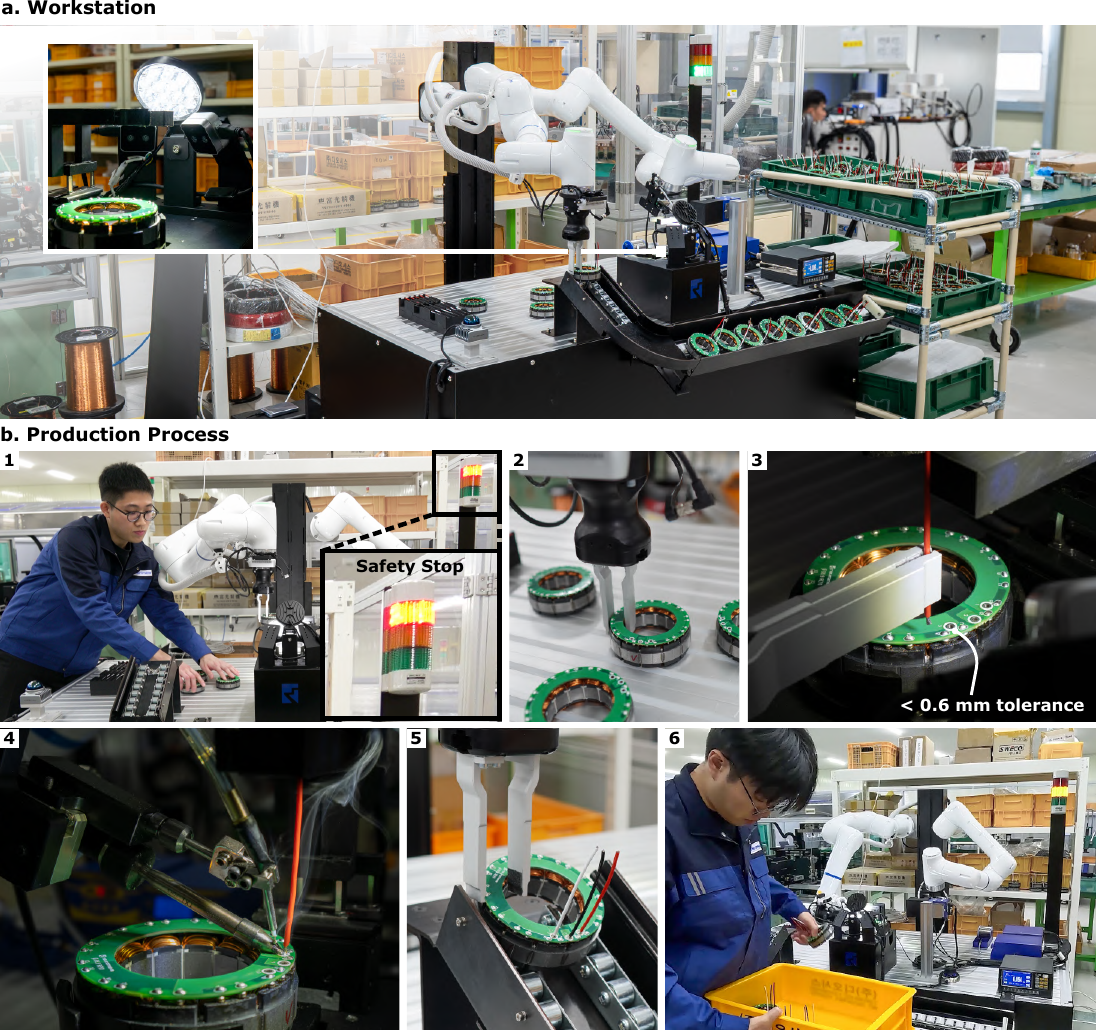}
\caption{\textbf{Production line deployment.}
\textbf{(a)} Overview of the workstation deployed on the factory floor.
\textbf{(b)} Production process and division of labor. A human worker loads motor cores and consumables into the cell (1). The robot then executes the cable insertion-and-soldering sequence, including motor placement/orientation, sequential insertion of the three-phase power cables, soldering, and tip cleaning (2–5). Finally, a human worker removes the completed motor and transfers it to the next stage of the line (6).}
\label{fig:production_overview}
\end{figure}

We conducted a factory-floor validation (Figure \ref{fig:production_overview}).
We integrated the proposed system within an existing automation cell on an electric-motor  production line and evaluated it through extended long-run operation ($\approx$ 5 h 10 min).
The process includes picking motors from random poses, inserting deformable cables, and performing soldering. The cable handling is especially challenging for rule-based automation because of the compliant materials and tight tolerances.
Over 108 consecutive motors with real components and consumables, the system achieved an average cycle time of \unit[159]{s} (average experienced-operator takt time $\approx$ \unit[141]{s}) and a 99.4\% success rate based on product-level \ac{QC} tests. This performance was achieved with less than 20 min of real-world data per task.

These results demonstrate the real-world viability of modular learning-augmented automation and establish a practical route to extending industrial automation to tasks that remain difficult for conventional methods.

\subsection*{Task Description and Challenges}

We target a manufacturing station on an electric-motor production line, namely three-phase power-cable insertion and soldering. The robot is required to execute the full station workflow, from picking a motor core from a table and placing it into the station to inserting and soldering the three cables, while meeting downstream product-level \ac{QC} requirements. Unlike laboratory test setups, this station must operate within an active production line, which imposes additional constraints on throughput, allowable iteration and data collection, and safe operation alongside human workers.

\begin{figure}[H]
\centering
\includegraphics[width=3.5 in]{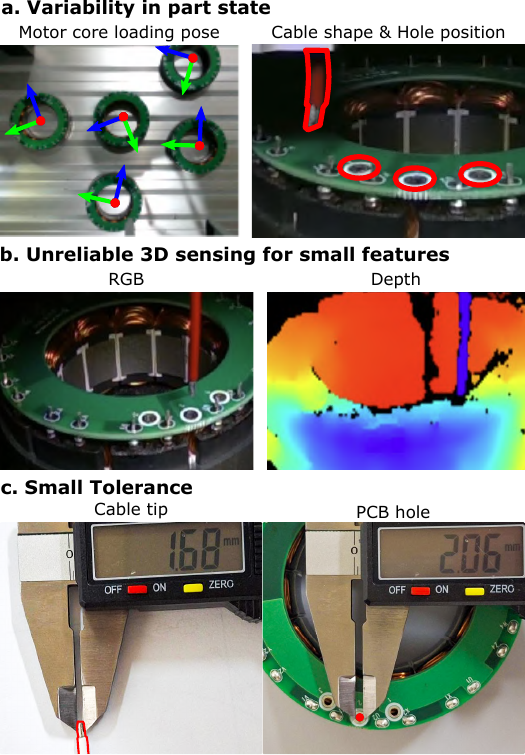}
\caption{\textbf{Task challenges.}
\textbf{(a)} Motor cores are loaded with random poses. PCB hole locations and cable shapes vary across specimens and trials.
\textbf{(b)} Unreliable depth sensing for small, thin, or reflective features near the insertion region.
\textbf{(c)}  Tight insertion clearance (0.3--0.6 mm) between the cable and PCB holes.
}
\label{fig:task_challenges}
\end{figure}

The motor cable soldering process has been challenging to automate with conventional waypoint-based approaches. First, the task involves geometric variation across parts (Figure \ref{fig:task_challenges}a). Motors are presented in random poses, hole positions can vary across instances, and cables deform during handling and insertion. Second, depth measurements are noisy and unreliable for thin or reflective structures (e.g., cables, holes, solder pads; Figure \ref{fig:task_challenges}b), making precise estimation of small-scale geometric features such as cable-tip pose and hole location difficult.
Third, insertion requires high precision under tight tolerances (Figure \ref{fig:task_challenges}c), as the clearance between the cable tip and the PCB hole is only 0.3 to \unit[0.6]{mm}.

The rest of the motor production line is already automated, with dedicated machines achieving stable throughput for operations such as coil winding, PCB dipping, and stator welding. In contrast, this insertion-and-soldering station has typically remained manual, performed by a skilled worker who solders the three connections and verifies quality in sequence.

\section*{Results}

\subsection*{System Overview}
We structured the automation pipeline as a sequence of modular manipulation tasks coordinated in a cyclic workflow (Figure \ref{fig:production_overview}b).

\begin{figure}[t]
\centering
\includegraphics[width=\textwidth]{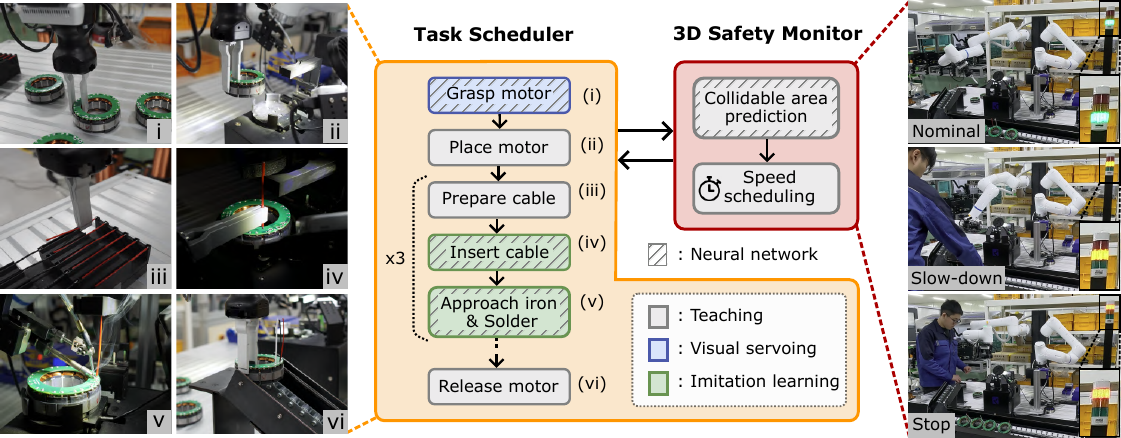}
\caption{\textbf{Learning-augmented robotic automation.} 
The software stack comprises modular task controllers and a safety monitoring module. Learned components are integrated at the task and safety levels to overcome the limitation of conventional automation.}
\label{fig:software_stack}
\end{figure}

\subsubsection*{Learning-Augmented Robotic Automation}

Our software stack is shown in Figure \ref{fig:software_stack}.
We extend classical industrial automation by retaining its core elements—an explicit Task Scheduler implemented as a \ac{FSM} and pre-taught motion segments for structured, repeatable parts of the workflow—while adding two learning-based capabilities that improve (i) task capability in unstructured, contact-rich subtasks and (ii) reactive collision prediction and risk reduction.

Learned controllers are integrated as modular primitives, implemented as callable modules with structured inputs/outputs, termination conditions, and explicit success signals that the \ac{FSM} uses for sequencing and fallback behaviors.
We developed two complementary categories: (i) Learned visual servoing controllers target kinematics-dominant subtasks in unstructured settings with substantial geometric variability by predicting a relative transformation from the current end-effector pose to the desired goal pose.
(ii) Imitation learning controllers address subtasks for which analytical modeling is difficult (e.g., deformable object handling and contact-rich interactions) by predicting action trajectories learned from human demonstrations.
This composition is conceptually related to Dreczkowski et al. ~\cite{dreczkowski2025learning}, which separates manipulation into alignment and interaction phases, although the criteria and implementation differ in our approach.

\begin{figure}[H]
\centering
\includegraphics[width=0.95\textwidth]{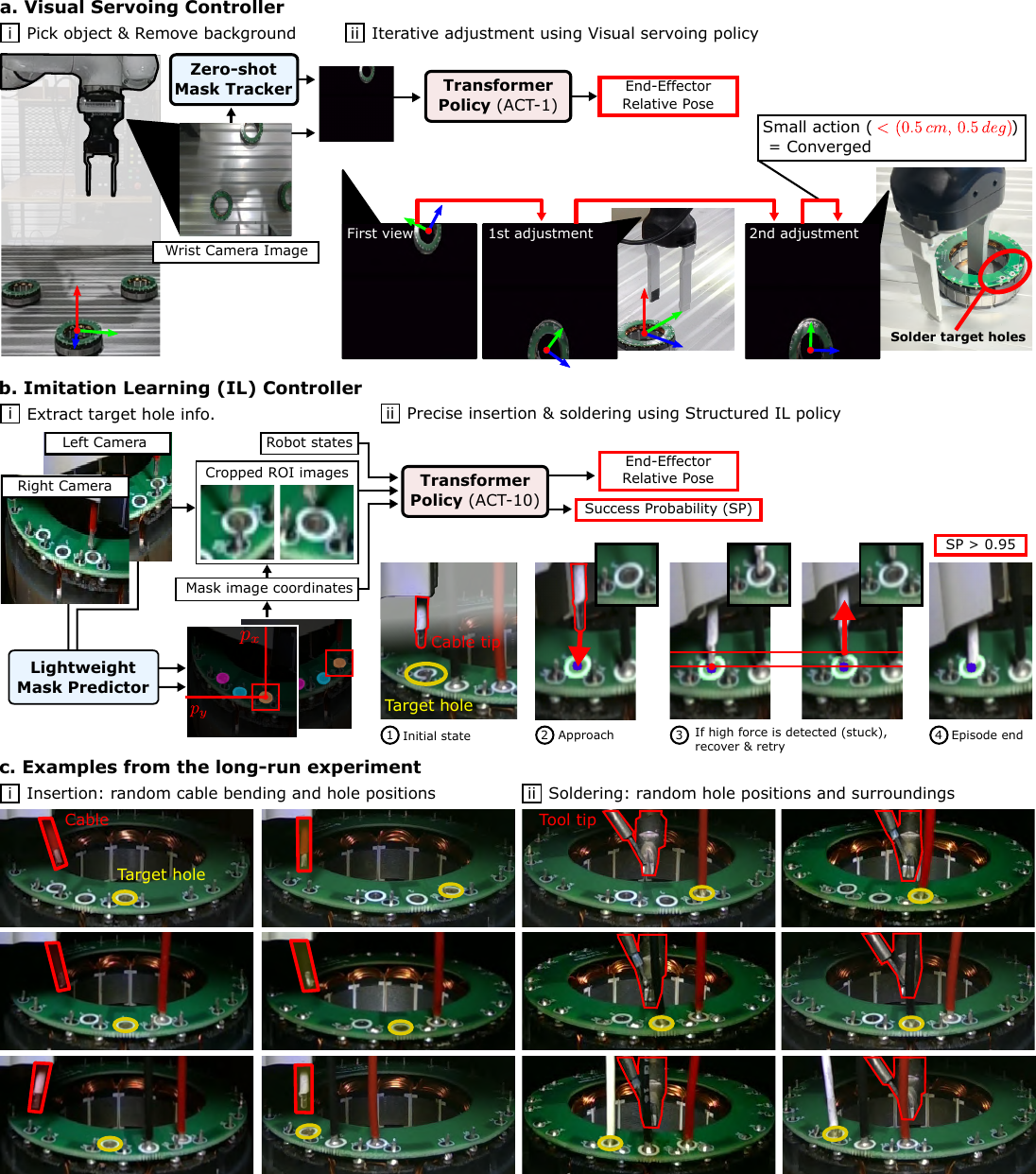}
\caption{\textbf{Behavior of learned controllers.}
\textbf{(a)} Visual servoing controller uses background-removed motor-core images as input (i) and iteratively adjusts the end-effector pose at each inference step until the target view is reached (ii).
\textbf{(b)} Imitation learning controller takes stereo RGB inputs from the worktable cameras. A lightweight mask predictor extracts the three PCB holes, and the policy is conditioned on the target-hole information (i). The target hole is specified by the high-level task scheduler. The end-effector is retracted by 2.5 to 4 mm if high force is detected by the load cell mounted below the motor jig (ii).
\textbf{(c)} Examples of task variability, including diverse cable configurations and target-hole locations.}
\label{fig:task_controller_qualitative}
\end{figure}

Learned controllers rely solely on RGB images for visual perception, unlike conventional automation that often depends on high-cost, high-precision 3D cameras. Rather than operating on full RGB images, we introduce task-specific structured observations as an inductive bias. For the learned visual servoing policy, a zero-shot mask tracker isolates the target object before predicting corrective motions (Figure \ref{fig:task_controller_qualitative}a-i). For the imitation learning policy, a lightweight mask predictor extracts a hole-invariant representation (ROI crops and target-hole location) that conditions the policy on the target-hole geometry (Figure \ref{fig:task_controller_qualitative}b-i). These inductive biases in the observation space reduce irrelevant visual variation, improving data efficiency and robustness under limited training data.

A 3D safety monitoring module is added to reduce collision risk when operating in populated workplaces. The module continuously monitors the workspace with a neural network that predicts obstacle occupancy from raw 3D point clouds~\cite{lee2025collisionRL, peng2020convocc}. The robot slows down when external objects are detected within the slowdown zone, reducing speed below conservative limits consistent with \ac{PFL}~\cite{ISO15066:2016} requirements. If obstacles are detected within the stop zone, the system triggers a protective stop~\cite{marvel2017SSM}. Zone details are provided in the Results section.

\subsubsection*{Hardware Setup}
The workstation consists of a bimanual collaborative robot built from two 6-\ac{DOF} arms (Supplementary Figure \ref{supp_fig:hardware_setup}). One arm is equipped with a two-finger gripper for motor and cable manipulation, while the other arm carries an automated soldering iron.

Three RGB-D cameras provide visual feedback; however, we use only the RGB streams (Supplementary Figure \ref{supp_fig:hardware_setup}-(1,2)).
A wrist-mounted camera on the gripper arm supports grasping parts from the table, while two fixed cameras positioned on the central worktable provide visual feedback during cable insertion and soldering.
A 3D LiDAR sensor is mounted at the workstation corner to monitor the surroundings (Supplementary Figure \ref{supp_fig:hardware_setup}-(3)).

The central worktable integrates a motor seat for insertion and soldering, along with a load cell (Supplementary Figure \ref{supp_fig:hardware_setup}-(4)). The load cell provides force feedback during insertion to detect cases where the cable becomes stuck. Such a failure case is difficult to distinguish from RGB images alone due to limited resolution (Supplementary Figure \ref{supp_fig:cable_stuck}). Upon detecting excessive load, the robot retracts the cable by a small distance (randomly sampled in \unit[2.5–4]{mm}) and retries the insertion.

Cables are loaded in a dedicated holder; an automated cable feeder could be integrated in future deployments (Supplementary Figure \ref{supp_fig:hardware_setup}-(5)).

\subsection*{Real-World Production-Line Validation}

\begin{figure}[H]
\centering
\includegraphics[width=5.0 in]{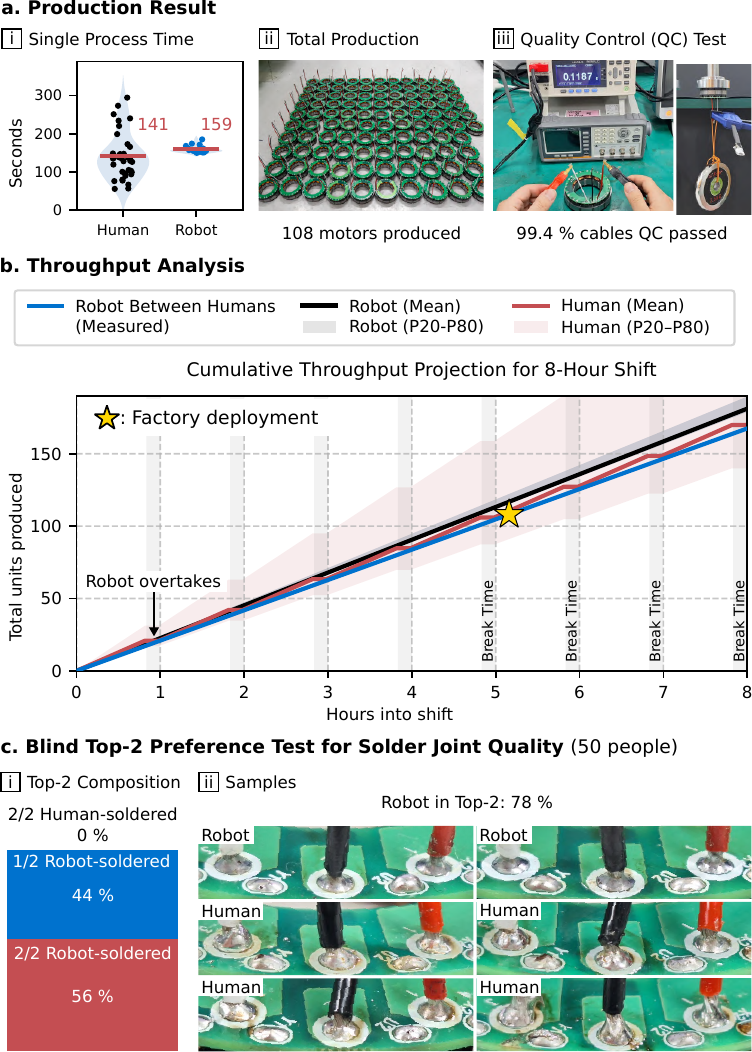}
\caption{\textbf{Production line deployment results}. 
\textbf{(a)} Production outcomes: \textbf{(a-i)} single-station cycle time (pick, insert, solder, tip clean) for a human worker vs. the robot. Human data comprise 37 measurements from three experienced workers; \textbf{(a-ii)} total output during the operation; \textbf{(a-iii)} downstream product-level \ac{QC} pass rate. 
\textbf{(b)} Cumulative throughput projection for an 8-hour shift. “Robot” is computed from the nominal cycle time, whereas “Robot between humans” uses the effective on-line takt time, including pauses for safety and material handling. 
\textbf{(c)} Blind Top-2 preference test of solder-joint quality ($\mathcal{N}=50$); \textbf{(c-i)} Top-2 composition and the fraction of robot results selected. \textbf{(c-ii)} Random samples used in the test.}
\label{fig:production_outcome}
\end{figure}

We validated the proposed learning-augmented automation through continuous operation on a live electric-motor production line at the Neuromeka Pohang factory (Figure \ref{fig:production_overview}, \href{https://youtu.be/M0e_vKzk7wk}{Supplementary Movie 2}), and summarized the key outcomes in Figure \ref{fig:task_controller_qualitative} and  Figure \ref{fig:production_outcome}. The deployment served as a stress test under production-line constraints, including sustained operation, safety for nearby human workers, and downstream product-level \ac{QC} requirements.

Figure \ref{fig:task_controller_qualitative} shows representative behaviors of the learned task controllers during the deployment (more examples in Supplementary Figure \ref{supp_fig:task_traj_vis}, \href{https://youtu.be/iuldg_uB5jU}{Supplementary Movie 4}).
Despite variability in part states, the system maintained stable long-run operation using learned, vision-based task controllers.  This was achieved with less than 20 minutes of real-world data per task. 
Table \ref{tab:data} summarizes the amount of data used in this study; approximately 8 minutes for motor grasping, 20 minutes for cable insertion, 4 minutes for soldering, and 9 minutes for training the PCB hole mask predictor.

Throughout deployment, the robots operated without physical fencing in a shared workspace. When workers approached to load materials or retrieve completed products, the system reduced speed or paused as needed and resumed autonomously (Figure \ref{fig:production_overview}b-1,6).

\begin{table}[H]
\centering
\small
\caption{\textbf{Training data size for each task.} For imitation learning controllers, additional data were collected using DAgger~\cite{ross2011dagger, kelly2019hgdagger} by rolling out the initially trained policy and gathering corrective demonstrations from failure states. The main failure modes addressed through DAgger were (i) encountering motor configurations not represented in the initial training data, causing the policy to fail to guide the cable or soldering-iron tip to the target hole, and (ii) the success predictor failing to recognize successful task completion. To address the latter, we collected additional demonstrations in which the cable remained fully inserted in the hole or the soldering iron tip remained at the target position.}
\label{tab:data}
\begin{tabularx}{0.85\linewidth}{X l c c c}
\hline
Task &  & Init & DAgger & \textbf{Total} \\
\hline\hline
\multirow{2}{*}{Motor grasping} 
    & Time           & 8 min 11 sec  & -      & \textbf{8 min 11 sec}  \\
    & \# of episodes & 8             & -      & \textbf{8}             \\
\hline
\multirow{2}{*}{Cable insertion}  
    & Time           & 19 min 18 sec & 16 sec & \textbf{19 min 34 sec} \\
    & \# of episodes & 180           & 5      & \textbf{185}           \\
\hline
\multirow{2}{*}{Soldering} 
    & Time           & 3 min 19 sec  & 38 sec & \textbf{3 min 57 sec}  \\
    & \# of episodes & 36            & 14     & \textbf{50}            \\
\hline
Hole mask predictor & Time & - & - & \textbf{9 min 7 sec} \\
\hline
\end{tabularx}
\end{table}

\subsubsection*{Production Performance and Quality}

The system produced 108 motors over approximately 5 h 10 min. The average nominal cycle time, which excludes safety pauses and slowdowns, was \unit[159]{s} per motor (Figure \ref{fig:production_outcome}a-i), which is approximately \unit[12.8]{\%} slower than the average experienced-operator takt time of \unit[141]{s} per unit.
However, the robot exhibited lower cycle-time variance than human operation (Figure \ref{fig:production_outcome}a-i), as human workers occasionally incur delays (e.g., exceeding \unit[3]{min}) when correcting insertion and soldering errors or due to operator fatigue.

Across the run, the robot executed 324 cable insertion-and-solder operations. Two operations failed when the insertion success detection logic triggered prematurely, leaving the cable insufficiently seated; the cable then disengaged during subsequent handling (\href{https://youtu.be/5JEBQnJQsW4}{Supplementary Movie 3}; more detailed analysis in Supplementary \ref{supp_sec:factory}). This yielded a success rate of \unit[99.4]{\%} per operation (322/324). Qualitative images of the soldered joints are provided in Supplementary Figure \ref{supp_fig:solder_joints}.

Processed motors passed the same downstream \ac{QC} checks used in routine production (Figure \ref{fig:production_outcome}a-ii,iii). The \ac{QC} protocol includes tensile and electrical tests. For tensile testing, a \unit[2.5]{kg} load was applied to the soldered cables for 1 min and the joint was inspected for disconnection or cracking. For electrical testing, resistance was measured between each pair of cables and verified to lie within the expected range for the motor’s intrinsic resistance.

\subsubsection*{Comparative Full-Shift Throughput Analysis}

Figure \ref{fig:production_outcome}b shows projected cumulative throughput over an 8-hour shift using three timing models.
To satisfy legal requirements in Korea, human work is organized into repeating 50-min work / 10-min break cycles. “Human” extrapolates the measured human takt time, and “Robot” extrapolates the robot’s nominal cycle time (both from Figure \ref{fig:production_outcome}a-i).
“Robot between humans” uses the effective on-line takt time, which includes pauses and slowdowns for worker access (material loading/removal).

Human cycle times show higher variance, as indicated by the P20–P80 interval in Figure \ref{fig:production_outcome}b (P20 and P80 denote the 20th and 80th percentiles, respectively).
Compared to the robot band, the human P20--P80 region is wider, reflecting occasional long-tail delays from interruptions, error recovery, and fatigue.
This implies that extrapolating from the mean takt time alone can underrepresent the variability that accumulates over an extended shift.
In addition, despite the robot’s slower nominal mean cycle time (Figure \ref{fig:production_outcome}a-i), the break-constrained human schedule reduces effective production time, so the “Robot” projection overtakes the “Human” projection after approximately \unit[1]{h} (Figure \ref{fig:production_outcome}b).

Beyond these timing projections, the shared-workspace setting yields labor-allocation benefits that are not captured by cycle-time analysis.
Because the robot does not require continuous human attention, the operator is mainly needed for periodic material loading and removal (approximately every 12--15 min).
During shared-workspace operation, the robot’s cycle time provided sufficient slack for the operator to perform post-soldering electrical \ac{QC} on completed motors (Figure \ref{fig:production_outcome}a-iii). This parallel work is not reflected in throughput metrics, but it can increase overall cell-level productivity by reallocating human effort without reducing robot utilization. A detailed productivity analysis is provided later.

\subsubsection*{Consistency of Solder-Joint Finishing}
In addition to throughput, the factory deployment showed improved production consistency.
Consistent solder-joint quality is also important for production-ready products. To evaluate the consistency of the solder joint (and its perceived visual quality), we conducted a blind Top-2 preference test (Figure \ref{fig:production_outcome}c).
Although not a rigorous reliability assay, this blind preference test mirrors real-world production inspection, where solder-joint appearance is routinely used as a first-line indicator of workmanship and consistency.

We randomly selected six finished motors: two robot-soldered samples and four human-soldered samples produced by different workers (Figure \ref{fig:production_outcome}c-ii). All six samples passed the QC test. Fifty participants with engineering and non-engineering backgrounds were shown the six samples and asked to select the two motors with the best-looking solder joints; participants were not informed whether each sample was robot- or human-soldered.
Details of the participants, the questionnaire, and the response distribution are provided in Supplementary \ref{supp_sec:blind_preference_test:design} and Supplementary Figure \ref{supp_fig:blind_test_details}.

Participants selected both robot-soldered motors in \unit[56]{\%} of trials, and selected one robot-soldered and one human-soldered motor in the remaining \unit[44]{\%}.
Across all Top-2 selections (50 participants $\times$ 2 choices), robot-soldered motors accounted for \unit[78]{\%} of votes, and the most frequent Top-2 pairings consistently included robot-soldered samples (Figure \ref{fig:production_outcome}c-i). A detailed statistical analysis, including significance testing, is provided in Supplementary \ref{supp_sec:blind_preference_test:statistic_analysis}. Together, these results suggest that, in addition to meeting QC requirements, our system produces solder-joint finishing that is visually competitive and consistent relative to the human baseline.

\subsubsection*{Changes in Production Workflow and Productivity}
To assess whether the demonstrated system performance translates into practical production gains, we analyzed how the system changes the production workflow and overall productivity. Before automation, a single operator was responsible for executing and monitoring the PCB dipping machine, manually conducting motor cable soldering, and performing \ac{QC} on completed motors. These three activities were performed sequentially because motor cable soldering required continuous operator involvement and accounted for more than half of the stage processing time. Specifically, the combined stage cycle was approximately \unit[319]{s}: \unit[72.6]{s} for PCB dipping, \unit[162.8]{s} for motor cable soldering, and \unit[83.6]{s} for \ac{QC}. Soldering thus accounted for approximately 51\% of the stage cycle. Its \unit[162.8]{s} duration includes associated operator movement and task-switching time. In contrast, the \unit[141]{s} reported in Figure \ref{fig:production_outcome}a-i represents the standalone soldering cycle, excluding this workflow overhead.

By automating motor cable soldering with the proposed robotic system, the operator no longer needs to remain continuously engaged in this production stage and instead intervenes periodically for material loading and unloading, machine recovery, and inspection of completed motors. This enables PCB dipping, robotic soldering, and QC inspection to proceed in parallel rather than sequentially. Based on the measured robot cycle time, the cycle time of this production stage decreases to 160–176 s (assuming a conservative 10\% scheduling margin), corresponding to an 81.3–99.4\% increase in stage throughput. Propagating this reduction through the complete manufacturing process, which requires 3196.6 s per motor, yields an estimated total production time of 3038–3054 s per motor, corresponding to a 4.7–5.2\% increase in overall production throughput.

Production anomalies and associated human interventions introduce only a small operational overhead. During the factory deployment, two cable insertion failures were detected during routine operator QC inspection and required manual correction. Each failure required approximately 47 s to repair, resulting in a total correction time of 94 s over 324 insertion-and-soldering operations. This additional intervention increased the overall production time of the 5 h 10 min deployment by approximately 0.5\%, indicating a negligible impact on overall production throughput. Supplementary \ref{supp_sec: anomaly_detection} and \href{https://youtu.be/KW-GrarzYkI}{Supplementary Movie 7} provide further details on potential production anomalies, their occurrence rates, detection methods, and recovery procedures. The anomaly cases include both those observed during factory deployment and plausible production-line disturbances identified during system development. Most anomalies were either not observed during deployment or required only brief operator intervention, resulting in little additional overhead under normal production conditions.

\subsection*{Comparison with Task Controller Variants} \label{sec:result_04}

We quantitatively compare our proposed task controllers against representative baseline methods for learning-based manipulation.

The evaluated baselines are as follows:
\begin{itemize}
    \item \textbf{Naive IL} uses the same \ac{ACT}-based imitation learning framework as our method~\cite{zhao2023ACT}, but removes our image-processing pipeline and structured visual features.
    The policy takes full-resolution RGB images as input and directly predicts actions. This baseline evaluates whether a generic visuomotor imitation learning formulation, without task-specific inductive bias in either the observation or action space, is sufficient for the target tasks.
    
    \item  \textbf{\ac{VLA}} employs a model with higher capacity and a stronger visual representation learned from large-scale pretraining. We use a fine-tuned $\pi_{0.5}$ model~\cite{intelligence2025pi0.5} for the corresponding tasks with full-resolution RGB images as input. This baseline is included to evaluate whether increased model size and pretrained representation alone can replace task-specific learning formulations and structured observations.

    \item \textbf{Conventional (Depth)} represents a traditional robotic automation pipeline based on explicit 3D perception and rule-based waypoint generation. It estimates the 3D positions of the cable tip and target PCB hole using image segmentation and depth sensing, and then executes predefined waypoint motions using the known camera pose and soldering tool geometry. For motor grasping, no depth-based baseline is provided because comparable grasping problems are commonly automated in industry using high-precision 3D sensing and CAD-based pose estimation \cite{pickit3d, cong2021visionReview}; we therefore did not reimplement the specialized perception and calibration pipelines.

    \item \textbf{Conventional (RGB)} represents a classical RGB-only visual servoing pipeline based on explicit feature extraction and hand-designed control. For motor grasping, it detects three PCB holes and aligns the resulting triangle to a target pose using a PD controller (Supplementary Figure \ref{supp_fig:classic_rgb_vs}a-i). For cable insertion, it detects the cable tip and target-hole image coordinates from stereo RGB images and computes robot motion using the standard image-based visual servoing (IBVS) formulation~\cite{chaumette2006classicRGBvs} with an interaction matrix calibrated from demonstration data (Supplementary Figure \ref{supp_fig:classic_rgb_vs}b-i). For soldering, no RGB baseline is provided because reliable feature detection is infeasible due to severe occlusion of the soldering tip by the already inserted cables.
\end{itemize}

All methods are trained on the same dataset where applicable, or with an identical data budget when modifications are necessary. Further implementation details are provided in Supplementary \ref{supp_sec:baseline}.

\begin{figure}[H]
\centering
\includegraphics[width=3.5 in]{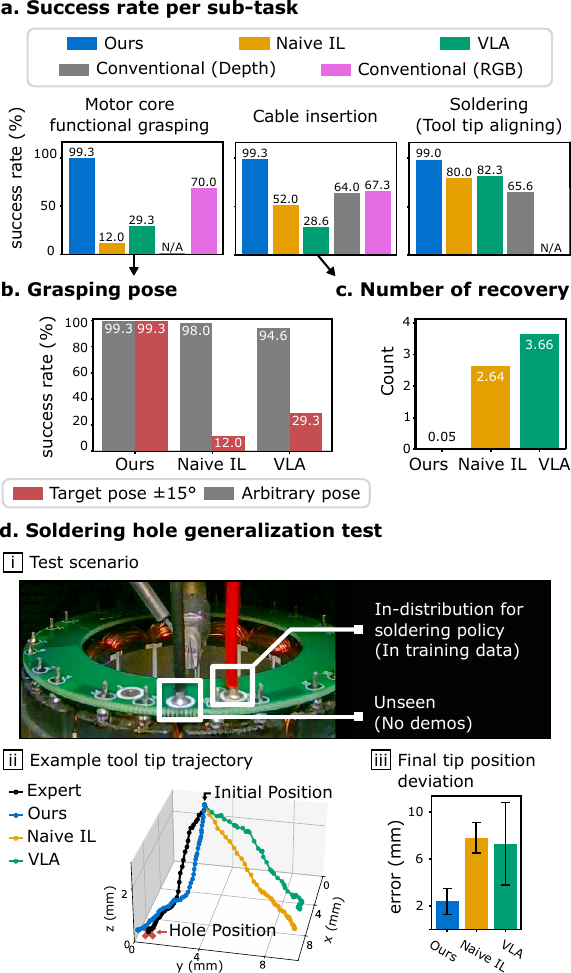}
\caption{\textbf{Comparison with task controller variants.}
\textbf{(a)} Sub-task success rates for motor-core functional grasping, cable insertion, and soldering (tool-tip alignment).
\textbf{(b)} Breakdown of grasping outcomes, separating overall grasp success from grasps achieved in the correct pose.
\textbf{(c)} Average number of recovery attempts during cable insertion.
\textbf{(d)} Soldering-hole generalization test.
\textbf{(d-i)} Test setup with an in-distribution hole (used in demonstrations) and an unseen hole (no demonstrations).
\textbf{(d-ii)} Example tool-tip trajectories from different policies.
\textbf{(d-iii)} Mean final tool-tip position error relative to expert demonstrations.}
\label{fig:task_controller_quantitative}
\end{figure}

\subsubsection*{Sub-task Performance Comparison}

Figure \ref{fig:task_controller_quantitative}a summarizes success rates for each unit task. All experiments use the same hardware setup for fair comparison. Each method is evaluated over 150 trials per task. For cable insertion and soldering, trials are evenly distributed across three PCB holes (50 per hole). For cable insertion, five different cables are used per hole to introduce variation. Additional evaluation details are provided in Supplementary \ref{supp_sec:task_eval}.

Our method achieves over \unit[99]{\%} success across all tasks, outperforming all baselines. Conventional baselines are limited by their reliance on explicit feature extraction to drive hand-designed control. The RGB-only baseline is vulnerable to PCB reflections, occlusions, and low image resolution, resulting in feature localization errors that lead to task failure (Supplementary Figure \ref{supp_fig:classic_rgb_vs}a-ii, \ref{supp_fig:classic_rgb_vs}b-ii). The depth-based baseline additionally suffers from noisy depth measurements, leading to inaccurate 3D pose estimation. While higher-precision industrial 3D vision solutions could mitigate this issue, they are typically costly.

Other learning-based baselines, including Naive IL and \ac{VLA}, achieve lower success rates than both our method and the conventional baseline on insertion under the same data budget. Large-scale pretraining of the \ac{VLA} model provides limited benefit for this industry-specific task. Despite stronger visual representations or larger model capacity, these methods do not consistently meet the task-specific precision requirements in this data-limited setting.

\subsubsection*{Data-Efficient Functional Grasping via Visual Servoing}

For motor grasping, the main difficulty is not reaching the motor but achieving the target grasp pose in which the three PCB holes appear in the wrist-camera view (Figure \ref{fig:task_controller_qualitative}a-ii). This orientation standardization is required to reduce downstream variability in hole locations for cable insertion and soldering.

As shown in Figure \ref{fig:task_controller_quantitative}b, all methods can robustly reach the motor and achieve physical contact. However, both Naive IL and \ac{VLA} often produce nearby but incorrect orientations. They tend to converge to a consistent visual configuration that is adequate for grasping, but does not satisfy the orientation requirement.

In contrast, we formulate motor grasping as a visual servoing problem. The policy predicts relative corrective motions in SE(3) toward the target view, rather than directly imitating demonstrated trajectories. This introduces an inductive bias in the action space: the controller is trained to reduce residual pose error with respect to the target view.

\subsubsection*{Structured Visual Observations Improve Robustness and Generalization}
For cable insertion and soldering, our method uses a hole-invariant policy that conditions actions on (i) the target hole’s image coordinates and (ii) cropped images around the hole, instead of the full RGB image used by Naive IL and \ac{VLA}. 

Figure \ref{fig:task_controller_quantitative}c shows that our method requires fewer recovery attempts during insertion than Naive IL and \ac{VLA}. The improvement compared to Naive IL baseline suggests that the structured, hole-centric visual inputs improve robustness by focusing the policy on the task-relevant features.
This localized observation reduces sensitivity to irrelevant visual regions, such as previously soldered cables, which often distract full-frame policies.

We further evaluate hole generalization for the soldering task (Figure \ref{fig:task_controller_quantitative}d).
We train the policy with only 10 demonstrations collected on a single hole, and then test on a different, unseen test specimen with one cable already inserted. The desired outcome is to solder a different hole under these conditions.

As shown in Figure \ref{fig:task_controller_quantitative}d-ii,iii, our soldering controller aligns the tool tip more closely with a previously unseen hole at test time than the other baselines. Naive IL and \ac{VLA} do not generalize to new holes without collecting new demonstrations for each hole, as they learn a direct mapping from full-scene images to actions without an explicit mechanism to retarget. In our experiments, language prompting did not change the pretrained \ac{VLA} behavior. 
While expected given this formulation, the implication is practical: instance-wise generalization enables reusing the same policy across targets, improving data efficiency for tasks that require repeated deployment across similar but distinct targets.

\subsection*{Safety Monitoring and Risk Reduction}
\label{sec:result_05}

\begin{figure}[H]
\centering
\includegraphics[width=5.0 in]{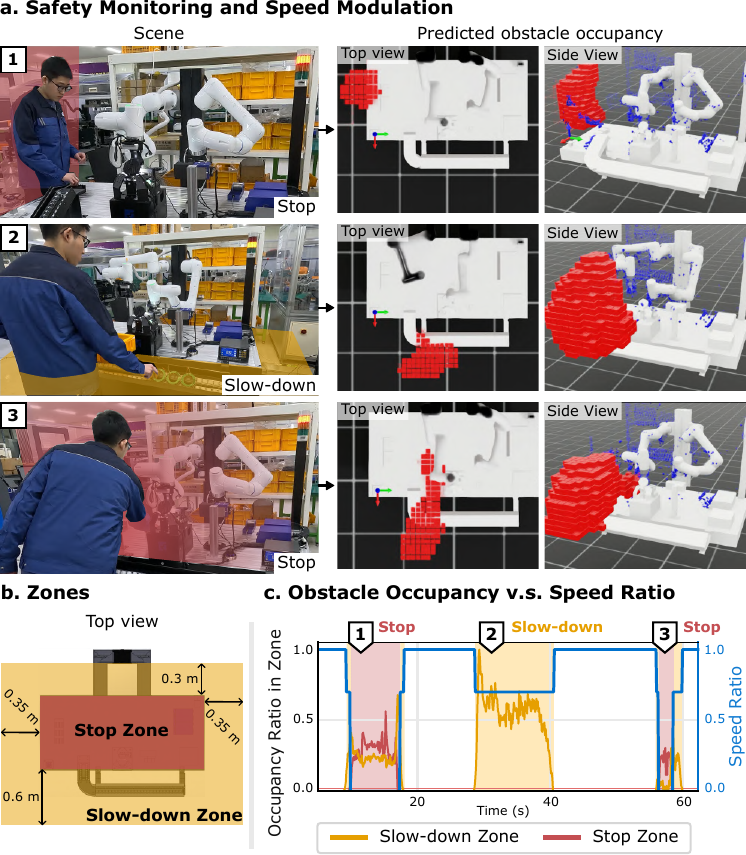}
\caption{\textbf{3D safety monitoring with speed modulation.}
\textbf{(a)} Example scenes (1--3) and the corresponding predicted obstacle occupancy (top/side views) from 3D point clouds; the controller commands {slow-down} or {stop} based on the occupied region.
\textbf{(b)} Top-view definition of the {slow-down} zone and the inner {stop} zone.
\textbf{(c)} Time series of obstacle occupancy ratio within each zone and the resulting commanded robot speed ratio; numbered intervals correspond to the scenes in \textbf{(a)}.}
\label{fig:speed_separation_monitor}
\end{figure}

Figure \ref{fig:speed_separation_monitor} illustrates how the safety monitoring system detects and responds to external objects. The current operation mode is communicated via a table-mounted LED (red: stop; yellow: slow-down; green: nominal). During the factory deployment, these interventions occurred repeatedly as workers approached the cell to load or unload materials (Figure \ref{fig:speed_separation_monitor}a, \href{https://youtu.be/qDf4CAxsT6M}{Supplementary Movie 5}).

As shown in Figure \ref{fig:speed_separation_monitor}b, we define two safety regions around the workstation: (i) a \textit{stop zone} covering the active workspace and (ii) a surrounding \textit{slow-down zone} extending 0.3--0.6~m beyond the table boundary. Based on voxelized point clouds, a neural network predicts obstacle occupancy within these regions (Figure \ref{fig:speed_separation_monitor}a, right). If the predicted occupancy ratio in a zone exceeds 0.1\%, the system triggers a protective response: a full stop if the stop zone is occupied, or a reduction to 70\% of nominal speed if occupancy is detected only in the slow-down zone (Figure \ref{fig:speed_separation_monitor}c).

Additional analysis of the safety monitor, including productivity trade-offs and compliance with \ac{PFL}~\cite{ISO15066:2016} standards, is provided in Supplementary \ref{supp_sec:safey_extra}.

The safety monitor is presented as a factory-validated research prototype rather than a certified industrial safety system. Formal certification is beyond the scope of this work and would require more rigorous validation against applicable industrial safety standards and certification requirements.

\section*{Discussion}

This work investigated whether recent learning-based approaches in manipulation can translate from laboratory demonstrations to reliable operation in real manufacturing. Our factory deployment results showed that structured integration of learned components can deliver stable long-run operation under production-line constraints.

Our system was grounded in existing industrial workflows—engineer-led setup, program-tree construction, and plug-and-play controllers—while operating under the key constraint of minimizing in-field data collection. Within this framework, we selectively introduced learned components at both the task and safety levels. This integration allowed data-driven modules to (i) expand automation capability under environmental and process variability, and (ii) support safe, efficient division of labor between human workers and robots.

Over a year of iterative development and on-site integration, we identified several design considerations that were critical for practical factory deployment. (See Supplementary Table~\ref{supp_tab:development_alternatives} and Supplementary~\ref{supp_sec:development_trials} for full details of the development process.)

First, explicit task decomposition into verifiable subtasks was critical for reliable industrial deployment. Automation had to be introduced incrementally and remain interpretable and debuggable, making purely end-to-end policies difficult to validate in practice. Accordingly, we adopted a modular design in which learning was applied only to subtasks requiring adaptation, while deterministic control was retained elsewhere. This hybrid structure helped satisfy practical requirements, namely high success rates and low cycle time, by constraining learning to well-defined subproblems and relying on fast, deterministic execution elsewhere.

Second, minimizing in-field data collection was essential. We addressed this by introducing task-specific inductive biases in both perception and control. These choices reduced learning complexity and enabled high performance under limited data budgets.

Finally, reducing avoidable downtime was important in human-shared workspaces. Manual triggers (e.g., push-buttons) or heuristic stop-and-go control~\cite{urSafety, marvel2017SSM, karagiannis2022adaptiveSSM} frequently fragmented workflows and reduced throughput. We addressed this with a neural 3D safety monitor that predicted potentially collidable regions from raw 3D point clouds in real time and modulated robot speed accordingly. This enabled tighter safety zones for speed modulation, allowing robots to adjust motion efficiently as workers approached and departed.

Looking ahead, an important challenge is extending the robot's operating domain to tasks that are harder to structure. Our current system assumes that automation processes can be decomposed into well-defined subtasks and that task-specific structured observations can be designed. Tasks that are difficult to decompose or lack clear visual structures may therefore require substantially more data. Developing methods that retain reliability while reducing data requirements in these less structured settings remains an important direction for future work.

Overall, our study illustrates a practical pathway toward more flexible and adaptive automation. Although we evaluated a soldering task, the design principles—modular task decomposition, task-specific inductive biases in both observation and action, and learned safety-aware coordination—are applicable to other tasks with geometric variability, difficult-to-handcraft behaviors, and imperfect fixturing. Supplementary \ref{supp_sec:additional_impl} provides qualitative implementation examples on a different hardware setup (\href{https://youtu.be/DJctgNjhxyc}{Supplementary Movie 8}) and task (\href{https://youtu.be/Te-zRx3jHxs}{Supplementary Movie 9}). Together, our results show the potential of learned modules to expand the automation frontier and unlock new classes of automatable tasks.

\begin{figure}[H]
\centering
\includegraphics[width=0.95 \textwidth]{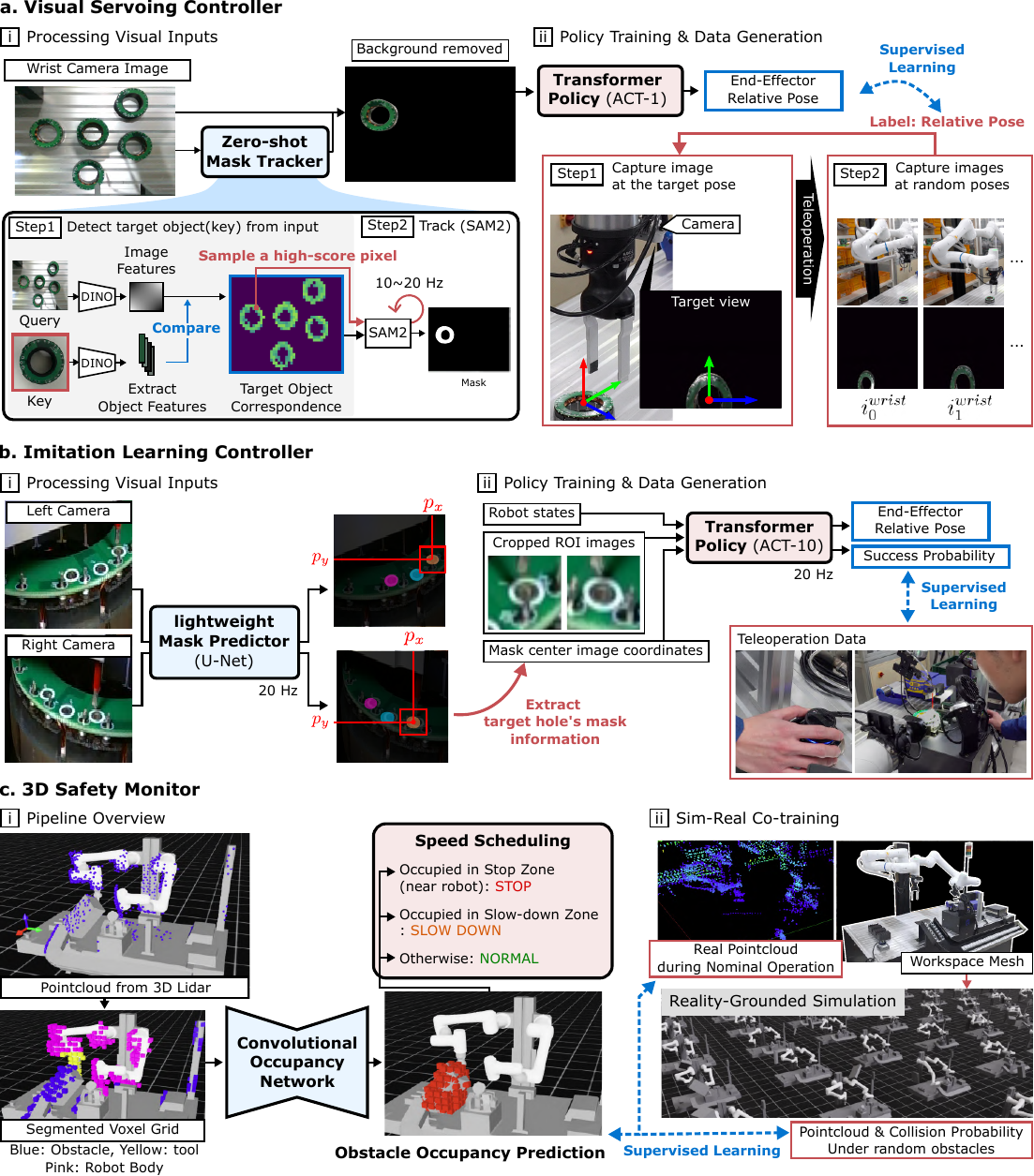}
\caption{\textbf{Overview of the learning-based components.}
\textbf{(a)} Visual servoing: \textbf{(a-i)} DINO-based key--query matching seeds a target mask, which is refined and tracked by SAM2 to create background-removed images; \textbf{(a-ii)} a transformer policy is trained on images collected at the target pose and at randomly perturbed poses.
\textbf{(b)} Imitation learning controller: \textbf{(b-i)} a lightweight U-Net predicts hole masks/IDs from stereo images; \textbf{(b-ii)} a transformer policy then processes target-hole information and robot states to output a relative pose action and a success probability. 
\textbf{(c)} 3D safety monitor: \textbf{(c-i)} a convolutional occupancy network predicts collidable regions from voxelized point clouds, and a speed scheduler switches between normal/slowdown/stop modes depending on the prediction; \textbf{(c-ii)} sim--real co-training builds a reality-grounded simulation for robust simulation training.
}
\label{fig:method}
\end{figure}

\section*{Methods}

Our system integrates learning at two specific levels while retaining a conventional industrial automation backbone: (i) task-level for perception-driven manipulation, and (ii) safety-level for reactive speed and separation monitoring. The remainder of this section details the design, training, and deployment of these learned components.

\subsection*{Learning-Based Task Controllers}

We employ two classes of learned controllers (Figure \ref{fig:method}a, \ref{fig:method}b)—visual servoing and imitation learning—each tailored to different task requirements.

\subsubsection*{Visual Servoing}\label{method:task_controller:vs}

For kinematics-dominant tasks where achieving the final end-effector pose is more critical than the specific trajectory to reach it, we employ a visual servoing approach (Figure \ref{fig:method}a). The visual servoing controller takes the current wrist-camera image $i^{wrist}_t$ as input and predicts a relative end-effector motion $a_{vs} \in SE(3)$, yielding an estimated target pose $s^{\prime}_f \in SE(3)$ computed as $s^{\prime}_f = s_t a_{vs}$, where $s_t \in SE(3)$ is the current robot pose, toward which the robot is driven. Rather than relying on explicit feature extraction (e.g., image keypoints, line detection, or CAD-based object 3D pose estimation) and heuristic motion generation, the controller is implemented as a neural network policy that directly infers corrective motions from visual observations~\cite{johns2021coarsetofine, yu2019siamese}. We use a masked RGB image as input, as further detailed in the “Preparing Observation” paragraph.

\paragraph{Task:}
We apply the visual servoing method to the motor grasping task (Figure \ref{fig:software_stack}-i). 
Given multiple motors randomly placed on a table with varying positions and orientations, the robot should grasp a single motor at a time in a desired configuration, where the three PCB holes face the wrist camera with a nearly consistent orientation (Figure \ref{fig:method}a-ii). This standardized grasp reduces variability in hole positions after pickup and simplifies downstream cable insertion and soldering. Since motors are always placed on a flat table, the motor grasping controller uses a top-down grasp with fixed end-effector roll and pitch, and the action space is defined as relative translations in $x$, $y$, and $z$ along with a relative $yaw$ rotation.

\paragraph{Preparing Observation:}
To scale the controller to scenes containing multiple motors and to encourage learning of task-relevant visual features, we operate on masked RGB images rather than raw images. In these masked observations, all regions except the target motor to be grasped are blacked out. A common approach to obtain such masks is to combine a segmentation model (e.g., SAM2~\cite{ravi2024sam2}) with an open-vocabulary object detector to provide an initial prompt (e.g., Grounding-DINO~\cite{liu2023groundingDINO}). In practice, however, this pipeline proved to be highly sensitive to detector performance and frequently failed in cluttered scenes, making reliable deployment difficult without extra fine-tuning.

To address these limitations, we adopt a zero-shot mask tracker that localizes and tracks target regions through feature matching in a pretrained visual representation space (Figure \ref{fig:method}a-i). Given a query image (the current RGB observation) and a key image containing a single motor, we extract dense visual features using a pretrained DINOv2 model~\cite{oquab2023dinov2} and compute patch-wise feature similarity over the query image. This produces a heatmap in which high responses correspond to regions visually similar to the motor. We apply contour filtering to select a coherent region from the heatmap and sample several points within it as prompts for SAM2. The resulting mask is subsequently tracked across frames, enabling stable masked observations for visual servoing in multi-motor scenes. Refer to Supplementary \ref{supp_sec:mask_extractor} for implementation details.

The zero-shot mask tracker is used only during policy deployment. For training data preparation, masked images are generated semi-automatically using SAM2 with manual point-prompting in the first frame to ensure data quality; this manual prompting takes less than a minute since the remaining frames are labeled automatically through mask tracking by SAM2.

\paragraph{Deployment:}
During deployment, the visual servoing controller operates in an iterative closed-loop manner. At the first step or whenever the previously estimated target pose is reached, the robot is commanded to move toward the newly predicted target pose. This process repeats until convergence, defined by the action norm falling below predefined thresholds ($\parallel \Delta pos \parallel < 0.5cm$, $\parallel \Delta rot \parallel < 0.5 deg$). The iterative refinement is necessary to compensate for partial visibility of PCB holes, visual artifacts from reflections and lighting, and residual regression errors in training. Once the policy converges, visual servoing terminates and the robot executes a predefined grasp by moving down to a fixed table height and closing the gripper.
 
\paragraph{Training:}
To train the neural network policy, we collect data through teleoperation (Figure \ref{fig:method}a-ii). For each episode, the motor is initially placed on the table with a random position and orientation. The robot is then manually positioned in a target grasp pose that is ready for motor pickup. Beginning from the target pose, the robot is teleoperated via a 3D SpaceMouse to randomly move around the motor, generating a range of perturbed states around the goal pose (\href{https://youtu.be/OtBzvnc1d-Y}{Supplementary Movie 6}). During teleoperation, we record the wrist-camera image $i^{wrist}_t$ and the corresponding end-effector pose $s_t \in SE(3)$ expressed in the robot base frame. Let $s_0$ denote the first recorded pose, which corresponds to the target grasp pose. For each recorded frame, the visual servoing action label is defined as the relative transformation $a_{vs} = s^{-1}_t s_0$. This results in training pairs $(i^{wrist}_t, a_{vs})$ that supervise the policy to infer corrective motions from visual observations. This tailored data collection strategy, in which the teleoperated trajectories and the recorded action labels are decoupled, is highly data-efficient for kinematics-dominant tasks. It avoids the need to collect full reaching trajectories for every motor configuration while still achieving strong performance with minimal training data (Figure \ref{fig:task_controller_quantitative}a).

We adopt Action Chunking with Transformers (ACT)~\cite{zhao2023ACT} as the policy backbone. The action chunk length is set to one, as the policy outputs a single relative target pose per step. The model is trained using an L1 loss to predict action labels.

\subsubsection*{Imitation Learning}\label{method:task_controller:il}

For tasks that require high precision or contact-sensitive interaction in semi-structured settings, we employ an imitation learning approach (Figure \ref{fig:method}b). Unlike kinematics-dominant tasks that can be handled by sparsely commanding target poses, these tasks require high-frequency adaptive motion based on continuous visual feedback. The imitation learning controller takes multi-camera images and robot state as input and predicts a short-horizon trajectory of relative end-effector motions $a_{il} \in SE(3)^K$, often referred to as an action chunk ($K$: chunk size). We adopt the relative end-effector trajectory representation introduced by Chi et al.~\cite{chi2024umi}.

To integrate the imitation learning controller as a modular component within the system, the task scheduler must determine when a task is complete in order to transition to the next step. For visual servoing, task completion can be detected using the action norm due to its convergent behavior, but this criterion does not apply to imitation learning. We therefore design the imitation learning policy to predict both control actions and a success probability.

\paragraph{Task:}
We apply the imitation learning formulation to the cable insertion and soldering task (Figure \ref{fig:software_stack}-iv,v). 
These tasks are performed three times for each motor because the motor is three-phase, containing three PCB holes (i.e., hole1, hole2, hole3) and three cables (i.e., red, black, and white). Although the motor grasping controller significantly reduces variability in hole positions after placement on the jig, residual positional uncertainty remains that the above two controllers must handle. In particular, the three PCB holes are distributed within an approximately 60° angular sector and a radial range of about 4 mm. 
Since the motors are placed on a flat jig, the action space of both controllers is defined as relative translations from the current pose in task-space ($\delta x, \delta y,\delta z$), with the robot end-effector orientation kept fixed.

\paragraph{Preparing Observation:}
Although raw RGB images are commonly used for imitation learning controllers~\cite{zhao2023ACT, fu2024mobileACT, chi2024diffusionpolicy, black2024pi0, intelligence2025pi0.5}, we empirically found that a structured observation space is particularly beneficial for achieving high performance with limited training data. We design a hole-invariant observation space by considering the characteristics of our task, where the robot performs conceptually the same operation at three PCB holes that appear visually different in raw RGB images.

We first train a lightweight U-Net–based~\cite{ronneberger2015unet} hole mask predictor on camera stream data with semi-automatic labeling via SAM2 (Figure \ref{fig:method}b-i). Given a target hole selected by the task scheduler and its predicted mask, we compute the hole's centroid pixel coordinates by averaging the coordinates of all the mask pixels. We also extract a local image crop of size 60×60 pixels centered at the hole. The resulting observation space consists of cropped images around the hole for each camera, the corresponding holes' pixel coordinates, and the robot’s relative end-effector position.

This structured observation design is inspired by Kim et al.~\cite{kim2021gaze} and is tailored for precise manipulation. The holes' pixel coordinates guide the robot toward the vicinity of the target hole, while the locally cropped image provides focused visual features for fine motion adaptation during cable insertion or soldering-iron alignment. By using adaptive cropping centered on the target hole, the controller remains robust to visual artifacts such as previously soldered cables or scorch marks on the PCB. In contrast, relying on full RGB images would require collecting substantially more data covering diverse visual artifacts to learn representations that consistently attend to the target hole region.

\paragraph{Deployment:}
During deployment, the imitation learning controller operates until the predicted success probability exceeds a threshold, at which point the task is considered complete. We use a fixed inference chunk size rather than temporal ensembling~\cite{zhao2023ACT}, as it was not critical in our setup. When the cable becomes stuck during insertion due to slight alignment errors, which is often difficult to identify from low-resolution cropped images (Supplementary Figure \ref{supp_fig:cable_stuck}), we rely on load readings from a load cell mounted beneath the motor jig. When excessive load is detected, the robot moves upward by a random offset and retries insertion.

\paragraph{Training:}
The neural network policy is trained using demonstration data collected through teleoperation (Figure \ref{fig:method}b-ii). For each episode, the motor is randomly initialized on the jig within the approximate range of positional variability introduced by the motor grasping controller. The robot is then teleoperated using a 3D SpaceMouse to perform the task, either inserting the cable or approaching the soldering iron tip (\href{https://youtu.be/OtBzvnc1d-Y}{Supplementary Movie 6}). The teleoperation is conducted at 20Hz, which matches the control frequency of the trained policy. During teleoperation, we record images from the left and right cameras mounted on the solder table, the robot’s current end-effector pose, and the control inputs from the teleoperation device. The recorded data are processed into training pairs ($i_t^{left}$, $i_t^{right}$, $p_t^{rel}$, $a_{il}$) to supervise the policy, where $p_t^{rel} \in \mathbb{R}^3$ denotes the end-effector position relative to the initial pose of the episode. 

Binary success labels are generated automatically from demonstrations, with the final portion of each trajectory labeled as successful. In practice, the last eight steps (0.4 s) are labeled as success to account for human reaction delay when stopping teleoperation recording.

The imitation learning policy uses the ACT model as its backbone. Hole image coordinates are treated as additional state inputs and concatenated with robot proprioception before being passed through a linear layer and transformer~\cite{vaswani2017transformer} encoder. On the output side, two transformer decoders are used: one to predict the action trajectory and the other to predict the success probability. The model is trained using an L1 loss for action prediction and a binary cross-entropy loss for success prediction.

An alternative approach commonly explored for high-precision tasks is real-world reinforcement learning~\cite{luo2024serl, luo2024hilserl}. While reinforcement learning could serve as another learning-based control module for automation in future work, we do not consider it in this study due to the difficulty of safe and efficient exploration in our target setting~\cite{li2026failure}. In our experience, real-world exploration in the task setup repeatedly damaged fragile components such as the soldering-iron tip and the cable tip. In contrast, imitation learning enables safe data collection through human demonstrations and allows explicit data quality control.

\subsection*{Learning-Based 3D Safety Monitor}\label{method:ssm}

In conventional automation, safe human–robot interaction is typically ensured by designing systems to comply with \ac{SSM} guidelines~\cite{ISO15066:2016}. These implementations commonly regulate robot motion using distance measurements from 2D laser scanners~\cite{urSafety, marvel2017SSM, karagiannis2022adaptiveSSM}. While effective in structured environments, such distance-based strategies are often overly conservative and difficult to deploy in tight human–shared workspaces, where occlusions and complex surrounding structures degrade measurement reliability. To address these limitations, we adopt a learning-based strategy that predicts collidable regions directly from raw 3D point clouds (Figure \ref{fig:method}c). The predicted collidable areas are then used to modulate robot speed between nominal, reduced, and protective stop modes based on predefined safety zones (Figure \ref{fig:speed_separation_monitor}b, Supplementary \ref{supp_sec:safey_extra:dynamic}). We enforce these joint speed limits using a model-based low-level controller~\cite{ko2025cbfqp, lee2022teleQP} to maintain safe operating speeds, complemented by the robot's intrinsic collision-stop mechanism~\cite{heo2019collisionDetection}. Furthermore, with an operational range of 90--120~cm, the workspace geometry inherently minimizes exposure to the head and face, avoiding the most stringent ISO/TS~15066 constraints~\cite{ISO15066:2016} (Supplementary~\ref{supp_sec:safey_extra:analysis}).

\subsubsection*{Collidable Area Prediction}

To avoid reliance on explicit online geometric pipelines—such as point matching~\cite{yang2025deepreactive}, mapping~\cite{oleynikova2017voxblox, sundaralingam2023curobo}, or handcrafted occlusion handling~\cite{zhu2020occlusion}—we formulate collidable area prediction with a single end-to-end neural network that maps point clouds to occupancy estimates. This design reduces system-level engineering complexity and scales efficiently to large workspaces.

\paragraph{Pipeline:}
Given a raw point cloud from the 3D LiDAR, we voxelize the data within the maximum region considered for collidable area prediction. Occupied voxels are then categorized into three classes—obstacle, robot, and end-effector tool—based on approximate cuboid geometries. For the robot, we construct a tree of cuboids that roughly cover the robot body, similar to the collision bodies defined in the robot’s URDF model. For the end-effector tool, we assume a single cuboid whose size is randomized during training and fixed during deployment according to the approximate gripper dimensions. The resulting voxel representation encodes each voxel by its spatial coordinate and a discrete label (0: empty, 1: obstacle, 2: robot, 3: tool). This entire process is implemented with {Warp}~\cite{warp2022} and computed in parallel on the GPU, enabling efficient real-time processing. We use a voxel size of 0.05 m, which is sufficient for safety monitoring.

The segmented voxel representation is passed to a neural network to predict collidable areas. We model this network as a Convolutional Occupancy Network~\cite{peng2020convocc}. It takes the segmented voxels together with the robot’s joint state as input and constructs a multi-plane feature field using convolutional layers. An implicit function, implemented as a multilayer perceptron (MLP), then predicts the occupancy probability at queried 3D coordinates conditioned on the extracted feature field. Because the model does not explicitly reconstruct the full 3D voxel grid~\cite{lee2025collisionRL}, it is memory efficient and scales well to large workspaces. In addition, occupancy queries for multiple 3D points can be evaluated in parallel on the GPU. During training, we subsample query points rather than evaluating all voxels. During deployment, we cache the voxel coordinates of the entire prediction area, query all points in batch, and classify a voxel as occupied if the predicted probability exceeds 0.5. In the soldering system, collidable area prediction runs in real time at 10 Hz.

\paragraph{Training:}
The neural network is trained using a combination of simulation data and real-world data. We employ a reality-grounded simulation in which each environment is constructed by spawning a 3D mesh of the real-world workspace obtained from CAD models. Although such meshes could also be acquired through real-world 3D scanning~\cite{torne2024rialto}, we leave this for future work. Robot joint configurations and motions in each environment are sampled around trajectories logged during real-world execution of the automation process to ensure realistic state distributions. Following our previous work~\cite{lee2025collisionRL}, each environment additionally includes randomly sized cuboids with randomized motion to simulate external obstacles, such as human workers, that are not present in the nominal setup. Raw 3D point clouds are generated by simulating a 3D LiDAR sensor with randomized mounting perturbations and additive gaussian noise to account for sensing uncertainty. The neural network is trained using a binary cross-entropy loss with ground-truth occupancy labels for external obstacles generated in simulation. We use IsaacLab~\cite{mittal2025isaaclab} for the simulation framework.

Due to unmodeled elements such as grasped motors, wired cables, robot tubing, and sensor latency, a model trained solely on simulation data exhibits limited sim-to-real transfer and tends to overestimate occupancy in regions without obstacles. To mitigate this issue, we additionally collect real-world LiDAR data while running the automation system without human presence (less than 8 minutes of data was recorded). Since these real-world data do not contain external obstacles, they are labeled as zero occupancy. The network is then co-trained using both simulation and real-world data, with each training batch composed of an equal proportion of samples from the two sources. This co-training strategy allows the network to learn from simulation data to handle occluded point clouds and external obstacles, while adapting to the unmodeled factors using real-world data.

\newpage

\section*{Acknowledgments}
We thank the Neuromeka executives and employees for supporting the experiments on the factory production line, participating in the blind preference test, and providing valuable feedback on the automation ecosystem. We also thank J. Kim, J. Kim, S. Kim, S. Lee, D. Ko, C. Kim, and I. Kwon for designing and setting up the robot workstation.

\section*{Author contributions}
Y.K formulated the main idea of the control and learning methods, implemented the system, and trained the learning-based modules. 
Q.N implemented the zero-shot mask tracker, \ac{VLA} training pipeline, and evaluated the suitability of reinforcement learning for the task.
Y.K and J.L refined the methodology, designed the experiments, and analyzed the data. 
Y.K and Q.N designed the hardware setup.
Y.K, Q.N, T.K, and J.L conducted experiments and prepared the manuscript.
Y.H provided insights from an industrial automation perspective, identified the necessity of the soldering automation, and supported preparation for long-run factory validation.

\section*{Funding statement}
This research was funded by the Industrial Technology Innovation Program (P0028404) of the Ministry of Trade, Industry and Resource of Korea.

\section*{Competing interests} 
The authors declare no competing interests.

\section*{Data availability}
Supplementary Data 1 contains the source data underlying Figure \ref{fig:production_outcome}, including individual human and robot cycle-time measurements and participant-level results from the blind Top-2 preference test. The model weights and training datasets are not publicly available because they are specific to the dedicated hardware setup and have limited standalone utility without it.

\section*{Code availability}
We have released the reusable software modules developed for this work in the Neuromeka-IL GitHub repository (\url{https://github.com/neuromeka-robotics/neuromeka-il}) under the Apache License 2.0, to support industrial practitioners and researchers developing learning-augmented robotic systems for other applications. The repository contains the complete code for the imitation learning controllers with the success-detection mechanism. Specifically, it includes code for data collection, training, deployment, and iterative improvement through DAgger~\cite{ross2011dagger, kelly2019hgdagger}. The repository also provides a \ac{FSM} example for constructing a hybrid system composed of multiple neural controllers and fixed-waypoint tracking controllers. The deployment interface is currently implemented for Neuromeka robots but can be adapted to other robotic platforms by replacing the corresponding robot communication functions.

\newpage

\bibliographystyle{Science}

\bibliography{scibib}

\newpage 

\section*{Description of Additional Supplementary Files}

\href{https://youtu.be/8Uu_UGyLyKY}{Supplementary Movie 1}. Paper summary\\
\href{https://youtu.be/M0e_vKzk7wk}{Supplementary Movie 2}. Deployment in a factory \\
\href{https://youtu.be/5JEBQnJQsW4}{Supplementary Movie 3}. Failure case in factory long-run\\
\href{https://youtu.be/iuldg_uB5jU}{Supplementary Movie 4}. Learning-based task controller evaluation\\
\href{https://youtu.be/qDf4CAxsT6M}{Supplementary Movie 5}. Learning-based 3D safety monitor evaluation\\
\href{https://youtu.be/OtBzvnc1d-Y}{Supplementary Movie 6}. Training data collection for task controllers\\
\href{https://youtu.be/KW-GrarzYkI}{Supplementary Movie 7}. Evaluation and stress testing of the anomaly monitoring framework\\
\href{https://youtu.be/DJctgNjhxyc}{Supplementary Movie 8}. Extension to a different hardware setup\\
\href{https://youtu.be/Te-zRx3jHxs}{Supplementary Movie 9}. Extension to a different task: chicken sauce brushing\\
Supplementary Data 1. Source data underlying Figure \ref{fig:production_outcome}\\

\newpage

\begin{titlepage}
\centering

\vspace*{2cm}

{\LARGE\bfseries Supplementary Information\par}

\vspace{0.8cm}

{\Large\bfseries
Learning-augmented robotic automation\\
for real-world manufacturing\par}

\vspace{1.5cm}

{\normalsize
Yunho Kim$^{*}$, Van Quan Nguyen, Taewhan Kim, Young Jin Heo, Joonho Lee$^{*}$\par}

\vspace{0.6cm}

{\normalsize
All authors are with Neuromeka Co., Ltd., Seoul, Republic of Korea\par}

\vspace{0.6cm}

{\small
$^{*}$Corresponding authors: yunho.kim@neuromeka.com, joonho.lee@neuromeka.com\par}

\vfill

\end{titlepage}

\setcounter{footnote}{0}

\setcounter{table}{0}
\makeatletter 
\renewcommand{\thetable}{\arabic{table}}
\captionsetup[table]{name=\textbf{Supplementary Table}}
\makeatother

\setcounter{figure}{0}
\makeatletter 
\renewcommand{\thefigure}{\arabic{figure}}
\captionsetup[figure]{name=Supplementary Figure}
\makeatother

\setcounter{subsection}{0}
\renewcommand{\thesubsection}{Note \arabic{subsection}}

\begin{figure}[H]
\centering
\includegraphics[width=\textwidth]{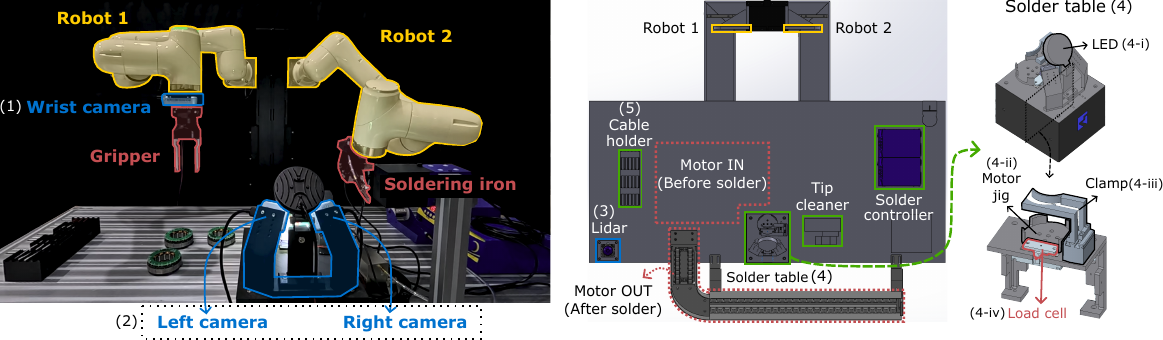}
\caption{\textbf{Hardware setup.} 
The workstation consists of bimanual collaborative robots, RGB cameras (1, 2), a 3D LiDAR (3), a central worktable (4), a cable holder (5), and soldering tools. The central worktable additionally integrates an LED light for stable illumination (4-i), a motor jig (4-ii), a motorized clamp (4-iii), and a load cell (4-iv). The clamp resolves geometric clearance issues caused by the bulky industrial gripper and soldering tool, thereby preventing collisions during operation.}
\label{supp_fig:hardware_setup}
\end{figure}

\begin{figure}[H]
\centering
\includegraphics[width=0.4\textwidth]{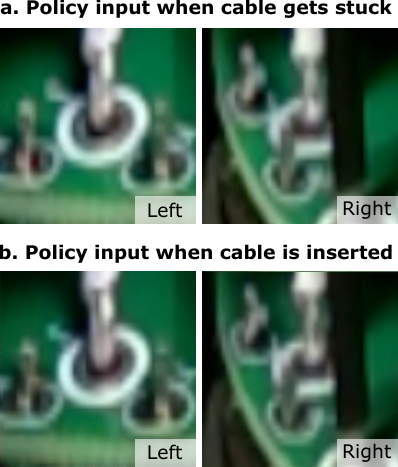}
\caption{\textbf{Comparison of visual observations during stuck and successful insertion.} 
\textbf{(a)} Policy inputs from the left and right cameras when the cable is stuck.
\textbf{(b)} Policy inputs when the cable is successfully inserted. When the cable is stuck, the load cell triggers a recovery procedure by retracting the cable and retrying insertion.}
\label{supp_fig:cable_stuck}
\end{figure}

\begin{figure}[H]
\centering
\includegraphics[width=\textwidth]{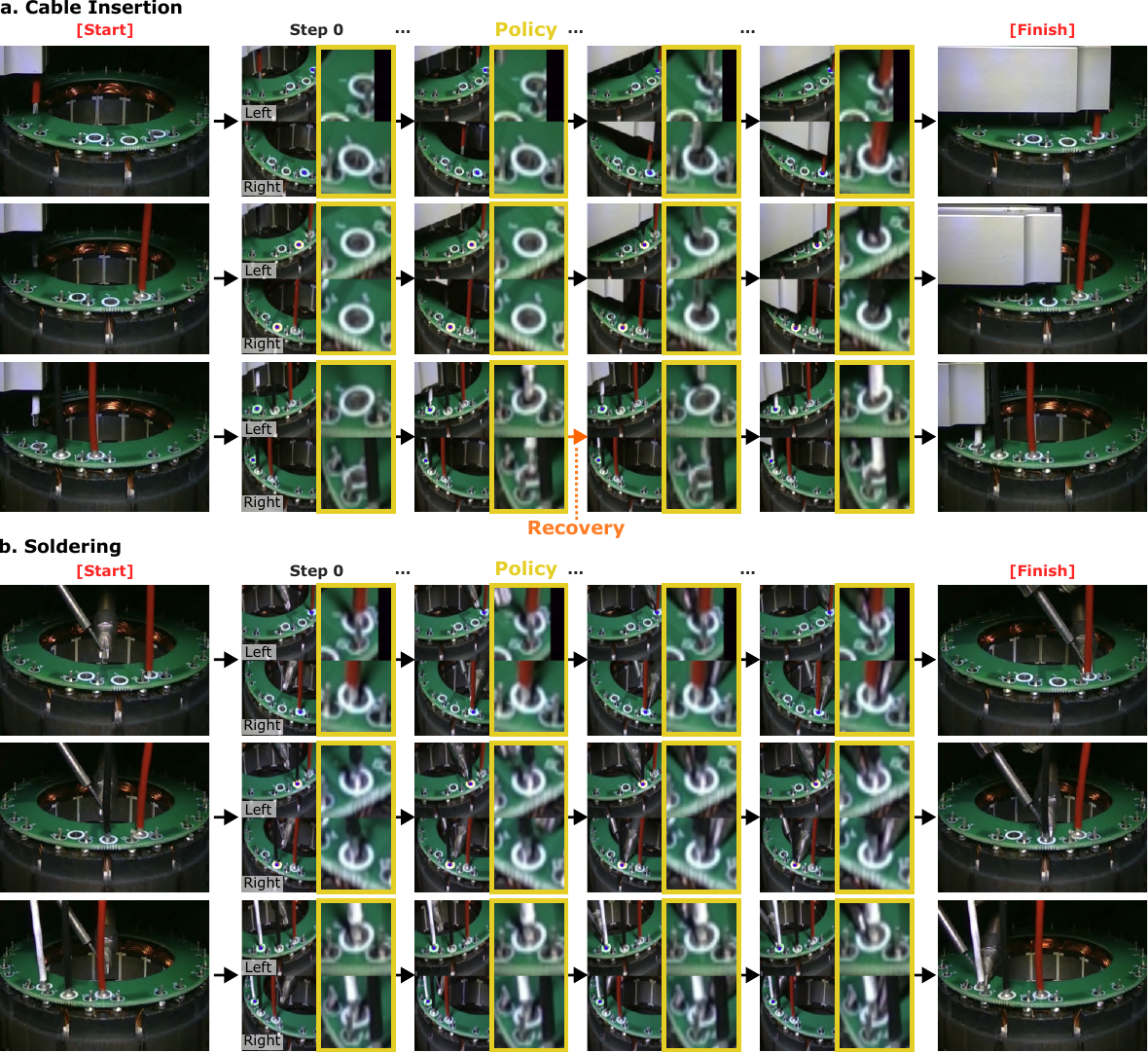}
\caption{\textbf{Observations of the imitation learning controller during rollouts.} 
\textbf{(a)} Cable insertion rollout.
\textbf{(b)} Soldering rollout. The controller uses image-space hole coordinates from the left and right cameras for coarse alignment. It performs fine adjustments when the cable or iron tip enters the cropped region. If the cable becomes stuck, a load cell triggers the robot to retract the cable and retry the insertion.
}
\label{supp_fig:task_traj_vis}
\end{figure}

\begin{figure}[H]
\centering
\includegraphics[width=\textwidth]{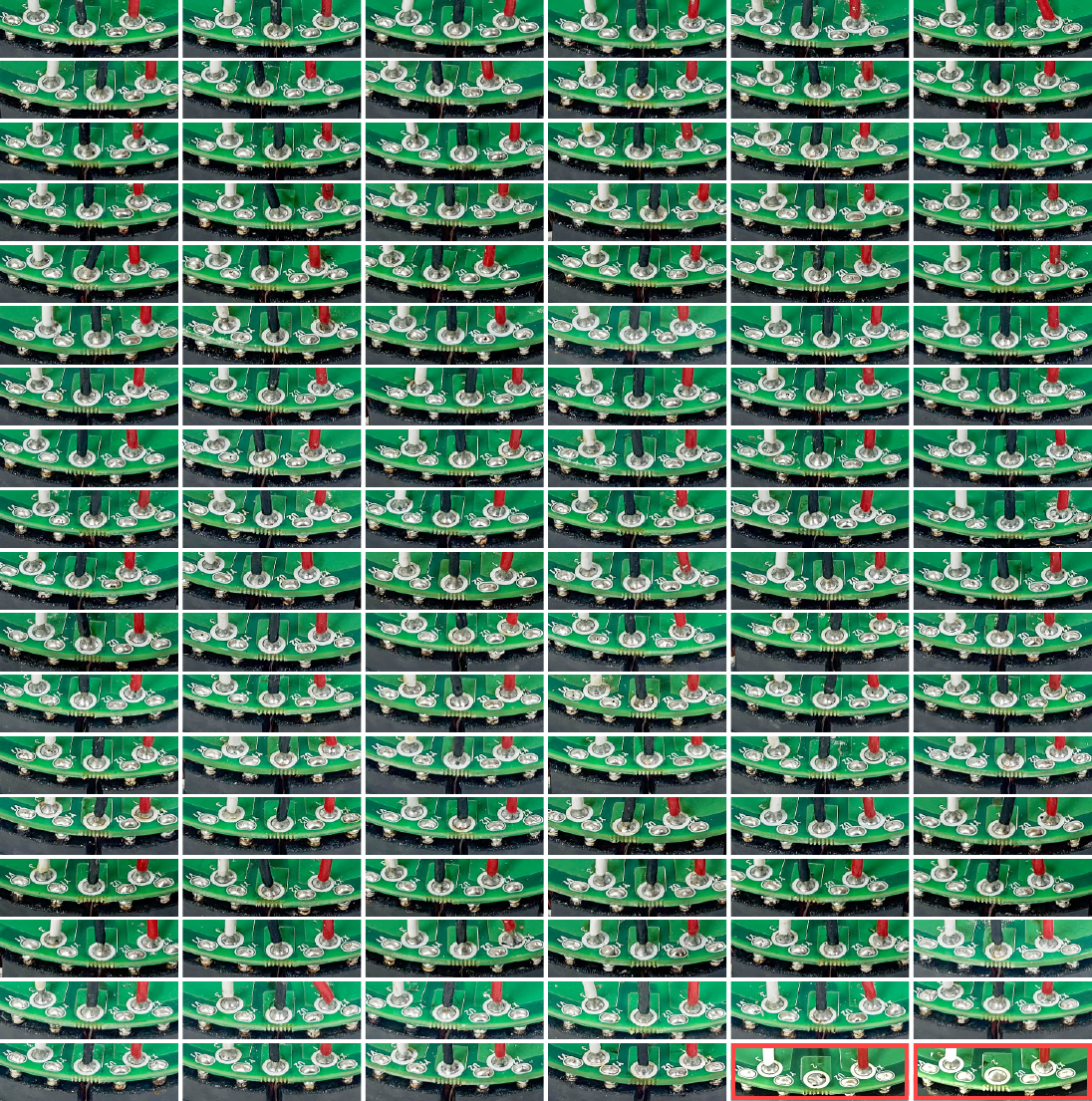}
\caption{\textbf{Soldered joints produced during factory deployment.} The last two images show the two failed soldered joints (black cable).}
\label{supp_fig:solder_joints}
\end{figure}

\begin{table}[H]
\centering
\fontsize{10pt}{11pt}\selectfont
\caption{\textbf{Development and design iterations for each component.} The iterative development process was guided by three design principles described in the Discussion and further clarified in Supplementary \ref{supp_sec:development_trials}.}
\label{supp_tab:development_alternatives}
\begin{tabularx}{\linewidth}{|>{\RaggedRight\arraybackslash}m{0.16\linewidth}|@{}X@{}|}
\hline
Component &
{\renewcommand{\tabularxcolumn}[1]{>{\RaggedRight\arraybackslash}m{#1}}\begin{tabularx}{\linewidth}{>{\Centering\arraybackslash}m{0.08\linewidth}|>{\RaggedRight\arraybackslash}m{0.30\linewidth}|X}
Iter. & Approach & Outcome
\end{tabularx}}
\tabularnewline \hline
Motor grasping controller &
{\renewcommand{\tabularxcolumn}[1]{>{\RaggedRight\arraybackslash}m{#1}}\begin{tabularx}{\linewidth}{>{\Centering\arraybackslash}m{0.08\linewidth}|>{\RaggedRight\arraybackslash}m{0.30\linewidth}|X}
1 & AnyGrasp~\cite{fang2023anygrasp} & AnyGrasp generates generic feasible grasps, whereas our task requires a precise grasp pose for the downstream subtasks. \\ \hline
2 & Naive IL & The policy was not sufficiently robust in scenes containing multiple motors. \\ \hline
3 & IL + masking & Robustness improved in scenes containing multiple motors. However, execution remained slow and success rate remained low. \\ \hline
4 & Learned visual servoing + masking (Ours) & Adopted as the final solution.
\end{tabularx}}
\tabularnewline \hline
Mask detector for motor grasping &
{\renewcommand{\tabularxcolumn}[1]{>{\RaggedRight\arraybackslash}m{#1}}\begin{tabularx}{\linewidth}{>{\Centering\arraybackslash}m{0.08\linewidth}|>{\RaggedRight\arraybackslash}m{0.30\linewidth}|X}
1 & Grounding-DINO~\cite{liu2023groundingDINO} & Motor detection was unreliable under illumination changes, PCB reflections, and clutter, resulting in missing or incorrect SAM2 prompts. \\ \hline
2 & Feature matching with DINOv2 (Ours)~\cite{oquab2023dinov2} & Adopted as the final solution.
\end{tabularx}}
\tabularnewline \hline
Cable insertion controller &
{\renewcommand{\tabularxcolumn}[1]{>{\RaggedRight\arraybackslash}m{#1}}\begin{tabularx}{\linewidth}{>{\Centering\arraybackslash}m{0.08\linewidth}|>{\RaggedRight\arraybackslash}m{0.30\linewidth}|X}
1 & Conventional & The quality of the depth images captured by the camera was insufficient for robust control. \\ \hline
2 & Real-world online RL~\cite{luo2024hilserl} & The policy frequently converged to suboptimal solutions. Moreover, safe and efficient exploration was difficult and required substantial human intervention during training. \\ \hline
3 & Naive IL & The insertion success rate remained low, and the policy was sensitive to distracting visual regions such as previously soldered cables. \\ \hline
4 & IL + cropping + hole coordinates & The policy became more robust to visual distractors, and the success rate improved. However, cables still often became stuck, which could not be detected by vision alone. \\ \hline
5 & IL + cropping + hole coordinates + recovery (Ours) & Adopted as the final solution.
\end{tabularx}}
\tabularnewline \hline
Iron approaching controller &
{\renewcommand{\tabularxcolumn}[1]{>{\RaggedRight\arraybackslash}m{#1}}\begin{tabularx}{\linewidth}{>{\Centering\arraybackslash}m{0.08\linewidth}|>{\RaggedRight\arraybackslash}m{0.30\linewidth}|X}
1 & Conventional & The quality of the depth images captured by the camera was insufficient for robust control. \\ \hline
2 & Real-world online RL~\cite{luo2024hilserl} & The policy frequently converged to suboptimal solutions. Moreover, safe and efficient exploration was difficult—particularly because the soldering-iron tip could be easily damaged—and required substantial human intervention during training. \\ \hline
3 & Naive IL & The success rate remained low, and the policy was sensitive to distracting visual regions such as previously soldered cables. \\ \hline
4 & IL + cropping + hole coordinates (Ours) & Adopted as the final solution.
\end{tabularx}}
\tabularnewline \hline
Safety monitoring system &
{\renewcommand{\tabularxcolumn}[1]{>{\RaggedRight\arraybackslash}m{#1}}\begin{tabularx}{\linewidth}{>{\Centering\arraybackslash}m{0.08\linewidth}|>{\RaggedRight\arraybackslash}m{0.30\linewidth}|X}
1 & Heuristic~\cite{marvel2017SSM,urSafety,karagiannis2022adaptiveSSM} & The approach was overly conservative and produced many false positives in the cluttered workspace. \\ \hline
2 & Learned perception (Ours) & Adopted as the final solution.
\end{tabularx}}
\tabularnewline \hline
\end{tabularx}
\end{table}

\subsection{Implementation Details for Zero-shot Mask Tracker}\label{supp_sec:mask_extractor}

To prepare the key for the zero-shot mask tracker, we use an image $x_k \in \mathbb{R}^{H_k \times W_k \times C}$ of the object (motor core) with a clean background, where $(H_k,W_k)$ is the image resolution after resizing to make both dimensions multiples of the DINOv2 patch size, and $C$ is the channel size. We then perform a forward pass of $x_k$ through DINOv2 to obtain the patch embeddings $y_k \in \mathbb{R}^{N \times D}$, where $N$ is the number of patches and $D$ is the embedding dimension. The background and foreground patches are separated by taking dot products of $y_k$ with a standard array \footnote{As provided by the DINOv2 project for background separation: https://dl.fbaipublicfiles.com/dinov2/arrays/standard.npy} and then thresholding them. The foreground patches are then combined into a 2D binary mask, followed by one round of binary erosion to remove patches that still contain background pixels. The final foreground embeddings $y_{kf} \in \mathbb{R}^{N_{kf} \times D}$, where $N_{kf}$ is the number of foreground patches, are saved for use at query time.

To query a mask for a new image $x_q \in \mathbb{R}^{H_q \times W_q \times C}$, we repeat the same procedure to obtain the query embeddings $y_{qf} \in \mathbb{R}^{N_{q f} \times D}$, where $N_{q f}$ is the number of foreground patches. We then calculate the similarity matrix
$$W_s = y_{q f}  y_{k f}^{T} \in \mathbb{R}^{N_{qf} \times N_{kf}}$$
and take the row-wise maximum:
$$s_i = \max_{j} (W_s)_{ij}, \qquad \mathbf{s} \in \mathbb{R}^{N_{qf}}$$
Finally, we threshold $\mathbf{s}$ to obtain the patch indices to sample from:
$$\mathcal{P} = \{ i : s_i \ge \tau \}$$

\subsection{Implementation Details for Baselines}\label{supp_sec:baseline}
All baselines are deployed and evaluated separately for each learning-based task: motor grasping, cable insertion, and soldering. For Naive IL and VLA, we did not train a single end-to-end policy for the complete motor-soldering workflow because teleoperating the entire process in a single episode required more than 5 minutes of continuous, high-precision human teleoperation. Such demonstrations were difficult to perform without intermediate pauses. Moreover, task-level evaluation provides a fairer comparison with our method and the conventional baselines, which are also implemented as task-specific controllers.

Naive IL and VLA baselines require task demonstrations for training. Accordingly, for "cable insertion" and "soldering", the same dataset as ours was used. For "motor grasping", since our method formulates the task as visual servoing and does not rely on demonstration trajectories, we collected demonstrations of equal size within the same motor pose randomization range.

Additional implementation details for VLA and Conventional (RGB) baselines are described below.

\subsubsection{VLA}
We finetune a single $\pi_{0.5}$ model~\cite{intelligence2025pi0.5} with different prompts for each task. The model accepts three RGB images $I_t^i$ for $i \in \{1, 2, 3\}$ corresponding to ``wrist image'', ``left image'', and ``right image'', a language task prompt $l_k$ where $k$ is the task index, and the robot's proprioceptive states $\mathbf{q}_t$. Supplementary Table \ref{supp_tab:vla_task_input_config} summarizes the exact camera input assignment and language prompt used for each task, which defines the modality and instruction context provided to the shared policy. The training code is adapted from https://github.com/Physical-Intelligence/openpi, and the finetuning hyperparameters are listed in Supplementary Table \ref{supp_tab:vla_hparams}.

\begin{table}[H]
\centering
\small
\caption{Input configuration for the \ac{VLA} baseline}
\label{supp_tab:vla_task_input_config}
\begin{tabularx}{\linewidth}{l c c c X}
\hline
Task & $I_t^1$ & $I_t^2$ & $I_t^3$ & $l_k$ \\
\hline
Motor grasping & wrist camera & 0 & 0 & ``grasp the motor'' \\
Cable insertion & 0 & left camera & right camera & ``insert the cable into the corresponding hole'' \\
Soldering & 0 & left camera & right camera & ``approach the hole and the cable'' \\
\hline
\end{tabularx}
\end{table}

\begin{table}[H]
\centering
\small
\caption{Hyperparameters for the \ac{VLA} baseline}
\label{supp_tab:vla_hparams}
\begin{tabular}{|l|l|c|}
\hline
\multirow{2}{*}{Train} & action horizon & 10 \\ \cline{2-3}
                       & batch size & 32 \\ \hline
\multirow{7}{*}{Scheduler} & type & cosine decay \\ \cline{2-3}
                           & warmup steps & 10000 \\ \cline{2-3}
                           & peak learning rate & $5e-5$ \\ \cline{2-3}
                           & decay steps & 1000000 \\ \cline{2-3}
                           & decay learning rate & $5e-5$ \\ \cline{2-3}
                           & ema decay & 0.999 \\ \cline{2-3}
                           & num train steps & 30000 \\ \hline
\end{tabular}
\end{table}

\subsubsection{Conventional (RGB)}
\begin{figure}[H]
\centering
\includegraphics[width=0.75\textwidth]{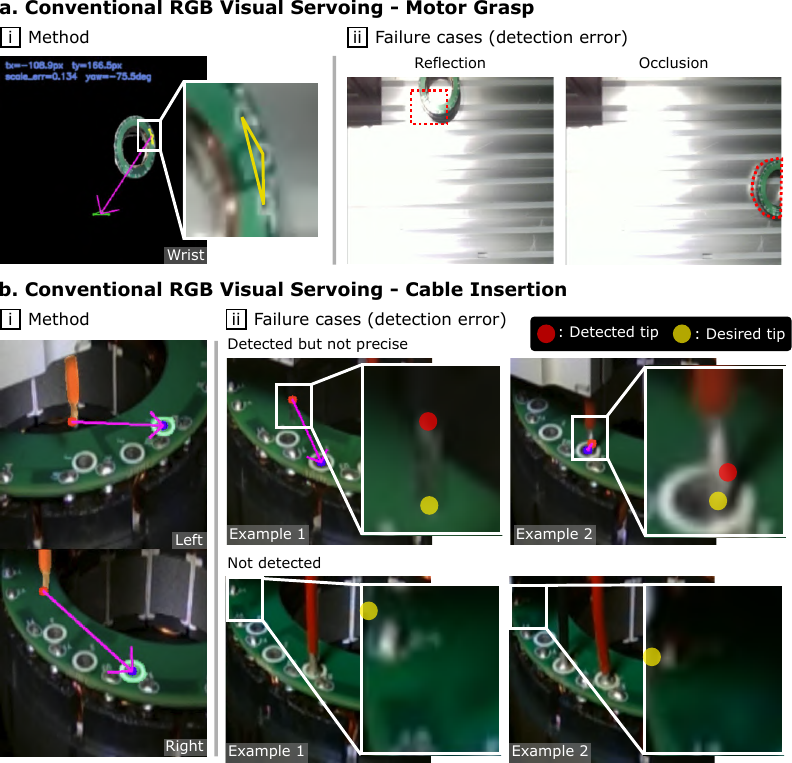}
\caption{\textbf{Conventional RGB-only methods and failure cases.}
\textbf{(a)} Conventional RGB visual servoing for motor grasping: \textbf{(a-i)} method based on detected PCB-hole geometry; \textbf{(a-ii)} representative failure cases caused by detection errors, including reflection and occlusion.
\textbf{(b)} Conventional RGB visual servoing for cable insertion: \textbf{(b-i)} image-based visual servoing (IBVS) using cable-tip and target-hole detections from the left and right cameras; \textbf{(b-ii)} representative failure cases caused by missed or imprecise feature detection.
}
\label{supp_fig:classic_rgb_vs}
\end{figure}

For motor grasping, we first train a lightweight mask predictor using the same dataset as our motor grasping controller. The predictor estimates the masks of the three PCB holes from an RGB image. From the predicted mask, the centroid of each hole is extracted to form a triangle representing the motor configuration in image space. A hand-designed PD controller then iteratively minimizes the error between the current and target triangles. Specifically, the image-space errors in the triangle centroid $(x, y)$, $scale$, and $orientation$ are mapped to relative end-effector translations in $x$, $y$, and $z$, and a relative $yaw$ rotation, respectively, using manually tuned proportional and derivative gains (Supplementary Figure \ref{supp_fig:classic_rgb_vs}a-i). The target triangle is obtained from the desired motor grasp configuration (Figure \ref{fig:task_controller_qualitative}a-ii).

For cable insertion, we first train a lightweight mask predictor using the same dataset as our hole mask predictor. The predictor estimates the masks of the cable and PCB holes from the left and right RGB images captured by two fixed cameras. From the predicted masks, the image coordinates of the cable tip and the centroid of the target PCB hole are extracted. Robot motion is then computed using the standard image-based visual servoing (IBVS) formulation~\cite{chaumette2006classicRGBvs}, where the relative end-effector translation is given by $\mathbf{u}=-\lambda(\mathbf{L}^{\top}\mathbf{L}+\mu\mathbf{I})^{-1}\mathbf{L}^{\top}\mathbf{e}$ (Supplementary Figure \ref{supp_fig:classic_rgb_vs}b-i). Here, $\mathbf{u}\in\mathbb{R}^{3}$ denotes the relative end-effector translation, $\mathbf{e}$ is the image-space error between the cable tip and the target hole for each camera, $\mathbf{L}$ is the interaction matrix, $\lambda$ is the controller gain, and $\mu$ is a damping coefficient. The interaction matrix is calibrated from demonstration data by fitting the relationship between image-space feature motion and robot end-effector motion. We also tested finite-difference calibration by perturbing the robot in Cartesian space, but demonstration-based calibration yielded better performance and was therefore adopted.

\subsection{Factory Validation Details and Failure Analysis}\label{supp_sec:factory}
The system was deployed on an electric-motor production line at the Neuromeka Pohang factory. During the long-run deployment, three temporary pauses occurred due to hardware-related or operational issues. The first was caused by solder material jamming in the feeder and required corrective maintenance. The other two occurred because the operator forgot to load cables into the holder. The deployed system did not include an anomaly detection mechanism to verify proper cable loading and respond to missing cables. Excluding the temporary stoppages, the net operational time was 5 hours and 10 minutes.

Two premature insertion-termination failures occurred during the factory deployment (\href{https://youtu.be/5JEBQnJQsW4}{Supplementary Movie 3}). The corresponding failure sequences are illustrated in Supplementary Figure ~\ref{supp_fig:failure_cases}. In both cases, the cable tip initially contacted the edge of the hole rather than entering it. The robot continued moving downward and slightly bent the cable until it suddenly snapped into the hole, causing both the cable and the robot end-effector to vibrate. During this vibration, the cable temporarily reached a fully inserted state and triggered the success detector. This caused the task scheduler (\ac{FSM}) to stop policy execution immediately. However, after the vibration subsided, the cable settled in a partially inserted state.

The system was subsequently improved by adding the anomaly-monitoring framework described in Supplementary \ref{supp_sec: anomaly_detection}. This enables automatic detection and response to the operational failures described above.

\begin{figure}[H]
\centering
\includegraphics[width=0.8\textwidth]{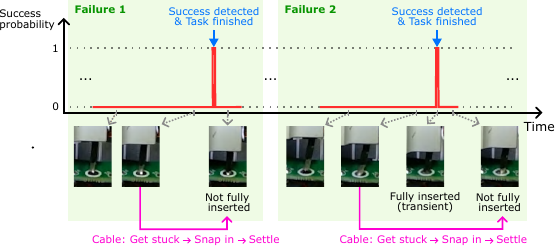}
\caption{\textbf{Failure cases encountered during factory deployment.} The y-axis shows the success probability predicted by the imitation learning controller.}
\label{supp_fig:failure_cases}
\end{figure}

\subsection{Blind Preference Test Details}\label{supp_sec:blind_preference_test}
\subsubsection{Test Design and Participants}\label{supp_sec:blind_preference_test:design}

\begin{figure}[H]
\centering
\includegraphics[width=0.85\textwidth]{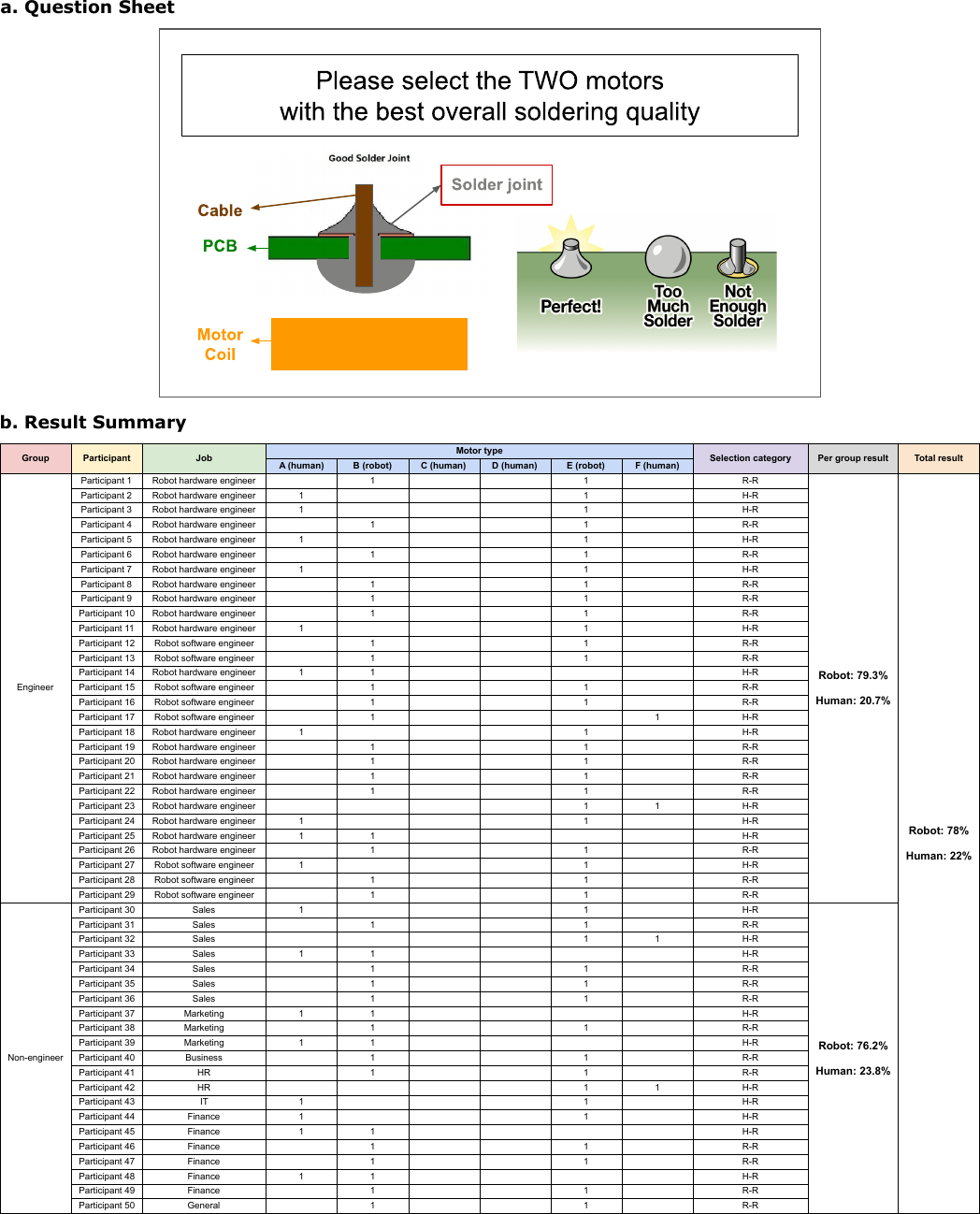}
\caption{\textbf{Details of the blind preference test.}
\textbf{(a)} Question sheet provided to the participants. 
\textbf{(b)} Summary of participant background, occupation, responses, and answer categories.}
\label{supp_fig:blind_test_details}
\end{figure}

The blind preference test was conducted with 50 randomly recruited participants, including both engineers and non-engineers to ensure diverse perspectives. Participants were asked to select the two motors with the highest perceived visual soldering quality. To support participants who were unfamiliar with cable soldering, a reference image illustrating an example of a properly soldered joint was provided in the questionnaire (Supplementary Figure \ref{supp_fig:blind_test_details}).

\subsubsection{Statistical Analysis}\label{supp_sec:blind_preference_test:statistic_analysis}
Among the 50 participants, 28 selected both robot-soldered samples and 22 selected one robot- and one human-soldered sample. Thus, robot-soldered samples accounted for 78 of the 100 individual selections.

To assess whether this result could reasonably arise by chance, we first derived the distribution expected under random selection. Let $X_i$ denote the number of robot-soldered samples selected by participant $i$. Because each participant selected two different samples without replacement, $X_i$ takes values 0, 1, and 2 with probabilities $6/15$, $8/15$, and $1/15$, respectively. Accordingly, the expected total number of robot-soldered selections across 50 participants is
\[
\mathbb{E}[T]
=
50\left(
0 \cdot \frac{6}{15}
+
1 \cdot \frac{8}{15}
+
2 \cdot \frac{1}{15}
\right)
=
33.3,
\]
where $T = \sum_{i=1}^{50} X_i$ denotes the total number of robot-soldered selections. The observed number of robot-soldered selections was 78, which is substantially higher than the expected count of 33.3 under random selection. 

For a more rigorous assessment, we performed an exact one-sided significance test to determine the probability of observing 78 or more robot-soldered selections under random selection. Let $a$, $b$, and $c$ denote the numbers of participants selecting two, one, and zero robot-soldered samples, respectively. Under the null hypothesis $H_0$ of random selection, the exact $p$-value is
\[
\begin{aligned}
p
&=
P(T \geq 78 \mid H_0) \\
&=
\sum_{\substack{
a+b+c=50 \\
2a+b \geq 78
}}
\frac{50!}{a!\,b!\,c!}
\left(\frac{1}{15}\right)^a
\left(\frac{8}{15}\right)^b
\left(\frac{6}{15}\right)^c \\
&=
4.14 \times 10^{-25}.
\end{aligned}
\]
This value is far below the conventional significance threshold of $p<0.05$, indicating that the robot-soldered samples were selected significantly more frequently than expected by chance. 

Overall, the analyses indicate that the observed result is highly unlikely to have arisen by chance and that it reflects a clear preference for the robot-soldered samples over the human-soldered samples.

\subsection{Task Evaluation Details}\label{supp_sec:task_eval}
Success criteria are defined per task. Motor grasping is considered successful if the motor is placed on the jig such that all three target holes fall within the load cell sensing region (±15° tolerance). Cable insertion is considered successful if the cable is fully inserted into the target hole. Soldering is considered successful if the iron tip reaches the target region around the cable–hole contact point. Additionally, for all tasks, a trial is counted as a failure if the task is not completed within a 20-second time limit.

\subsection{Extended Safety Monitor Analysis and Dynamic Stop Zone}\label{supp_sec:safey_extra}

This section provides three additional analyses: productivity metrics, a safety evaluation based on energy-related impact severity, and an extension of the system to adaptive workspace monitoring.

\subsubsection{Productivity Trade-off}\label{supp_sec:safey_extra:productivity}

Safety mechanisms inevitably reduce throughput when they slow or stop the robot. We quantified this trade-off by comparing our approach against a no-intervention upper bound (i.e., operation without speed modulation). We recorded raw LiDAR point clouds and robot motions during the production of 10 motors with workers present, then replayed this data offline to simulate different safety strategies and accumulate the resulting downtime.

Our method incurred an estimated 9\% increase in total production time relative to the no-intervention upper bound. For comparison, we evaluated a conventional distance-threshold baseline reflecting common industrial practice~\cite{urSafety, marvel2017SSM, karagiannis2022adaptiveSSM}, which triggers an immediate stop whenever any point-cloud return is detected within a 0.2~m fixed margin around the table (selected to envelop the protective distance derived via \ac{SSM} in Supplementary~\ref{supp_sec:safey_extra:dynamic}). This conservative strategy incurred a 23 \% production-time increase, demonstrating the overhead of heuristic stop-and-go protocols.

While full safety certification is outside the scope of this work, these results suggest that leveraging 3D information can support tighter, context-aware safety zones that reduce unnecessary stops and enable more fluid human--robot coordination in shared workplaces.

\subsubsection{Collision Risk Analysis within Slow Down Zone}\label{supp_sec:safey_extra:analysis}

To evaluate the safety of the deployed robot motion with respect to Power and Force Limiting (PFL) standards~\cite{ISO15066:2016}, we analyzed the trajectories using the SARA (Safe Autonomous human-robot collaboration through Reachability Analysis) framework proposed by Thumm et al.~\cite{thumm2024general}. The SARA framework provides mathematically conservative safety guarantees by directly bounding the robot's total operational-space kinetic energy ($T_r$). While this safety bound is applicable to collision with any body part, we specifically computed the operational-space inertia matrix and Cartesian velocity for the end-effector~\cite{haddadin2012making}, as it represents the most critical impact point due to its high velocity and exposure.

Considering the geometric constraints, our safety analysis focuses on the body regions most likely to be exposed to the robot: the hands, arms, and torso. The robot is mounted such that its end-effector operates strictly above a height of $0.9$~m, effectively mitigating the likelihood of collision with the lower body. Furthermore, the task design restricts the robot's operational volume to below the average head height ($<1.2$~m), rendering accidental contact with the head and face unlikely.

Supplementary Table \ref{supp_tab:sara_safety_utilization} shows that under worst-case unconstrained blunt contact, the peak kinetic energy of $0.295$~J stays well below the corresponding SARA thresholds for the hand ($0.49$~J), arms ($1.30$--$1.50$~J), and torso ($1.60$~J). Our analysis demonstrates that the system is consistent with PFL requirements for these reasonably foreseeable collision scenarios.

Although head and face contact was not included in the evaluated exposure scenarios, the corresponding SARA threshold can be further satisfied by reducing the slow-down speed from 70\% of the nominal speed, as used during the factory deployment, to approximately 43\%. This adjustment only requires changing the low-level speed limit and does not require retraining of the safety monitor.

\begin{table}[h]
\centering
\caption{Safety Verification using SARA Shield. The \textbf{Safety Ratio} ($T_r / T_{limit}$) indicates the percentage of the safety threshold used. Values $>1.0$ indicate a violation.}
\label{supp_tab:sara_safety_utilization}
\begin{tabular}{lc|cc|cc}
\hline
\multirow{2}{*}{\textbf{Body Region}} & \textbf{Limit} & \multicolumn{2}{c|}{\textbf{Right arm} ($T_r = 0.295$ J)} & \multicolumn{2}{c}{\textbf{Left arm} ($T_r = 0.167$ J)} \\ \cline{3-6} 
 & \textbf{[J]} \cite{thumm2024general} & \textbf{Ratio} & \textbf{Status} & \textbf{Ratio} & \textbf{Status} \\ \hline
Head (Face) & 0.11 & \textbf{2.68} & \textbf{Unsafe} & \textbf{1.52} & \textbf{Unsafe} \\
Hand & 0.49 & 0.60 & Safe & 0.34 & Safe \\
Lower Arm & 1.30 & 0.23 & Safe & 0.13 & Safe \\
Upper Arm & 1.50 & 0.20 & Safe & 0.11 & Safe \\
Torso (Chest) & 1.60 & 0.18 & Safe & 0.10 & Safe \\ \hline
\end{tabular}
\end{table}

\subsubsection{Dynamic Stop Zone}\label{supp_sec:safey_extra:dynamic}
For robots with large workspaces, a static stop zone covering the entire operational volume can be unnecessarily conservative, leading to frequent and inefficient stops.
To mitigate this, we implemented an adaptive protective distance approach, where the stop zone is represented as a set of dynamic spheres with variable radii centered at each of the robot’s link positions.

Following the Speed and Separation Monitoring (SSM) guidelines in ISO/TS 15066~\cite{ISO15066:2016, marvel2017SSM}, the protective distance (stop zone radius) $S$ is computed at every monitoring cycle ($\Delta t = 0.1$~s) as:
\begin{equation}
    S = v_h (t_r + t_s) + v_r t_r + b + C + z_r + z_s
\end{equation}
where $v_h$ and $v_r$ denote human and robot body velocities, $t_r$ is the system reaction time, $t_s$ is the stopping time, $b$ is the braking distance, $C$ is the intrusion distance (safety margin), and $z_r, z_s$ represent robot and sensor position uncertainties, respectively.

As illustrated in Supplementary Figure \ref{supp_fig:dynamice_stop_zone_viz}, this formulation allows each sphere to expand as the corresponding link velocity increases and contract as the robot slows down, effectively maintaining safety while maximizing operational efficiency.

Using parameter sets defined for our platform (Supplementary Table \ref{supp_tab:safety_monitor_hparams}), safety monitoring based on the dynamic stop-zone formulation resulted in an 11\% increase in total production time relative to the no-intervention upper bound. This remains more efficient than a conventional distance-based speed-modulation baseline. Future iterations will apply this dynamic zoning to more complex environments.

\begin{figure}[H]
\centering
\includegraphics[width=0.75\textwidth]{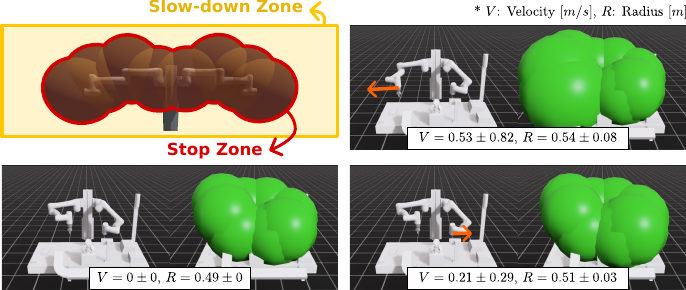}
\caption{\textbf{Dynamic stop zone.} Stop zone is represented as spheres attached to robot’s links. Sphere radii expand as robot’s speed increases and contract as it slows down. Arrows indicate robot’s direction of motion.}
\label{supp_fig:dynamice_stop_zone_viz}
\end{figure}

\subsection{Anomaly Monitoring for Long-Term Deployment}\label{supp_sec: anomaly_detection}
Long-term deployment of the robotic soldering system in production requires an anomaly monitoring module capable of managing failures beyond nominal operating conditions. In this section, we present an example anomaly monitoring framework to illustrate how such functionality can be integrated into the proposed automation system. This framework was implemented after the reported 108-motor factory deployment and was not active during that run.

The framework covers eight representative production anomalies, including those encountered during the factory deployment (Supplementary Table~\ref{supp_tab:anomaly_cases}). Anomaly detectors and recovery rules are implemented at the relevant workflow stages, triggering either automatic recovery or targeted operator intervention. For example, a foreign-object anomaly handler is incorporated into the motor-grasping stage, while insertion-related anomaly handlers are integrated into the cable-insertion stage. This extension was enabled by the modular architecture of the proposed system, which allows new task-local safeguards to be added without retraining the learned controllers or redesigning the overall automation sequence. 

\newcolumntype{L}[1]{>{\RaggedRight\arraybackslash}p{#1}}
\newcolumntype{C}[1]{>{\Centering\arraybackslash}p{#1}}
\newcolumntype{Y}{>{\RaggedRight\arraybackslash}X}

\begin{table}[H]
\begin{threeparttable}
\centering
\scriptsize
\caption{Summary of the anomaly monitoring framework. Potential anomaly cases, detection methods, recovery strategies, occurrence rates, and evaluation results are listed.}
\label{supp_tab:anomaly_cases}
\setlength{\tabcolsep}{4pt}
\renewcommand{\arraystretch}{1.15}
\begin{tabularx}{\textwidth}{|L{1.45cm}|C{1.45cm}|Y|C{1.15cm}|Y|C{0.85cm}|C{0.85cm}|C{1.1cm}|C{1.15cm}|}
\hline
\multicolumn{1}{|c|}{\textbf{Category}} & \multicolumn{1}{c|}{\textbf{Case}} & \multicolumn{1}{c|}{\textbf{Description}} & \multicolumn{1}{c|}{\textbf{Method}} & \multicolumn{1}{c|}{\textbf{Recovery}} & \multicolumn{1}{c|}{\textbf{OR}} & \multicolumn{1}{c|}{\textbf{DR}} & \multicolumn{1}{c|}{\textbf{RT [s]}} & \multicolumn{1}{c|}{\textbf{CT [s]}} \\ \hline
\multirow{2}{=}{Motor loading} & Incorrect motor size & A motor of an incorrect size is loaded. & Bbox & Automatic stop $\rightarrow$ Remove wrong motor $\rightarrow$ Resume & 0\% & 100\% & 8 & 0 \\ \cline{2-9}
& Foreign object & An unexpected object or improperly oriented motor is present in the workspace. & VLM & Automatic stop $\rightarrow$ Remove/correct object $\rightarrow$ Resume & 0\% & 98.3\% & 8 & 0 \\ \hline
Cable loading & Missing or incorrect cable & The cable is missing or an incorrect cable type is loaded. & Feature matching & Automatic stop $\rightarrow$ Replace cable $\rightarrow$ Resume & 0.6\% & 96.7\% & 16 & 0 \\ \hline
\multirow{5}{=}{Task execution} & Worktable lighting anomaly & The worktable LED is off or degraded, or the camera view is occluded. & Brightness & Automatic stop $\rightarrow$ Check LED/camera $\rightarrow$ Resume & 0\% & 100\% & 4 & 0 \\ \cline{2-9}
& Premature termination (insertion) & The insertion task terminates before a stable, fully inserted state is confirmed. & Classifier + Recheck loop & Automatic insertion retry & 0.6\% & 100\% & 0 & 0 \\ \cline{2-9}
& Insertion failure & The cable tip is too large for the PCB hole, or the observed configuration is OOD for the learned policy. & Task time limit & Automatic cable discard and insertion retry & 0.9\% & 100\% & 0 & 0 \\ \cline{2-9}
& Soldering failure & The soldering-iron tip is degraded, or contact among the tip, cable, and PCB hole is insufficient. & Classifier & Automatic stop $\rightarrow$ Inspect cause $\rightarrow$ (Repair solder tip $\rightarrow$) Resume/Restart $\rightarrow$ Correct affected motors & 0\% & 78\% & 4$\sim$180 & 55 \\ \cline{2-9}
& Solder feeder error & The feeder fails to supply solder wire or stops due to another malfunction. & I/O signal & Automatic stop $\rightarrow$ Repair feeder $\rightarrow$ Restart $\rightarrow$ Correct affected motors & 0.3\% & 100\% & 310 & 47 \\ \hline
\end{tabularx}

\begin{tablenotes}[flushleft]
\scriptsize
\item * \textbf{OR}: occurrence rate during factory deployment. \textbf{DR}: detection rate measured in subsequent stress tests. \textbf{RT}: mean time to recover production. \textbf{CT}: mean correction time per affected hole after production.
\item * Anomaly monitoring framework is complemented by periodic human inspection while unloading finished motors from the release track every 12–15 minutes.
\end{tablenotes}

\end{threeparttable}
\end{table}

\begin{figure}[H]
\centering
\includegraphics[width=0.95\textwidth]{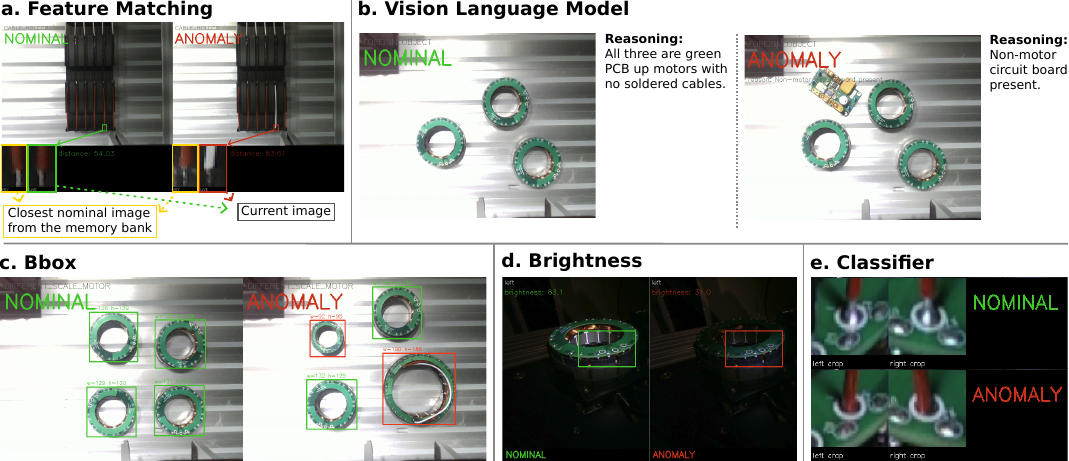}
\caption{\textbf{Anomaly detection methods.}
\textbf{(a)} Feature-matching-based detection of missing or incorrect cables. 
\textbf{(b)} Vision-language-model-based detection of foreign objects or incorrectly oriented motors. 
\textbf{(c)} Bounding-box-based detection of incorrect motor size. 
\textbf{(d)} Brightness-based detection of worktable lighting anomalies.
\textbf{(e)} Classifier-based detection of soldering failures.
}
\label{supp_fig:anomaly_detector_method}
\end{figure}

The implemented anomaly detection methods for each case are described below.

\begin{itemize}
\item \textbf{Incorrect motor size.}
The zero-shot mask tracker outputs an object mask for each detected motor. A bounding box is computed from each mask, and the motor is classified as anomalous if its width or height falls outside the nominal range (Supplementary Figure \ref{supp_fig:anomaly_detector_method}c).
\item \textbf{Foreign object.} \footnote{Although handled as a potential anomaly, foreign objects are not a critical anomaly because they generally do not disrupt operation. Their features differ from those of the motor key image, so the zero-shot mask tracker does not select them as tracking targets.}
A vision-language model receives the nominal image, current image, and a task-specific prompt asking whether foreign objects or incorrectly oriented motors are present. The model outputs both a binary anomaly decision and a short textual explanation (Supplementary Figure \ref{supp_fig:anomaly_detector_method}b). We use Kimi-K2.5~\cite{team2026kimi} model, running on a server within the factory’s local network, with the following system prompt. To avoid increasing cycle time, the model is queried with a wrist-camera image captured while the robot waits to unload the soldered motor, rather than immediately before motor grasping. 
\begin{promptbox}
\footnotesize
\textbf{System Prompt}\\
You are a visual quality inspector.\\
\\
The first image shows a nominal motor for reference. \\
The nominal motor has:\\
- Green PCB side facing upward\\
- Empty PCB holes — no cables soldered\\
\\
Look at the second image and answer:\\
- Is every visible object in the second image a motor?\\
- Flag ANOMALY if there is an object that is clearly not a motor (e.g. a screw, a different component type, package bag).\\
- Flag ANOMALY if any motor is flipped upside-down, i.e. the copper coil side is facing upward instead of the green PCB.\\
- Flag ANOMALY if any motor has cables already soldered into its PCB holes.\\
- Do NOT flag based on motor size or scale — that is handled separately.\\
- Do NOT treat bright light reflection or glare on the PCB surface as a broken motor, foreign object, soldered cable, or flipped motor.\\
- Partially visible motors (cut off by the image border) do not count as anomalies.\\
\\
Respond in exactly this format:\\
VERDICT: $<$normal or anomaly$>$\\
REASONING: $<$one or two sentences explaining your decision$>$\\
\end{promptbox}
\item \textbf{Missing or incorrect cable.}
Feature matching is performed following Roth \textit{et al.}~\cite{roth2022anomalyFM}. Feature embeddings extracted from the current image are compared against a precomputed memory bank of nominal images in which all cables are correctly loaded. An anomaly is detected when the nearest-neighbor feature distance exceeds a predefined threshold (Supplementary Figure \ref{supp_fig:anomaly_detector_method}a). We use concatenated WideResNet~\cite{zagoruyko2016wideRN} and DINOv2~\cite{oquab2023dinov2} features.
\item \textbf{Worktable lighting anomaly.}
The average image brightness is monitored, and an anomaly is detected when it falls outside the nominal range established during normal operation (Supplementary Figure \ref{supp_fig:anomaly_detector_method}d).
\item \textbf{Premature termination during insertion.}
To prevent transient success detection from terminating the insertion controller before the cable reaches a stable, fully inserted state, the insertion success signal is continuously rechecked for \unit[0.5]{s}. The insertion task is considered complete only if the success signal remains continuously valid throughout this period; otherwise, the controller continues execution until stable success is detected.
\item \textbf{Insertion failure.}
If insertion is not completed within a predefined timeout, the robot automatically discards the cable, grasps a new cable, and retries the insertion. This recovery strategy was employed during the factory deployment.
\item \textbf{Soldering failure.}
The insertion success detector is reused to detect soldering failures. After soldering, the central worktable clamp is released and the insertion success detector is evaluated. If the cable is no longer detected as fully inserted, the soldering operation is considered successful, as the cable remains fixed to the PCB by the solder joint. Conversely, if the detector still predicts a fully inserted cable, it indicates that the cable was not soldered and fell back into the PCB hole after the clamp was released. In this case, the operation is classified as anomalous (Supplementary Figure \ref{supp_fig:anomaly_detector_method}e).
\item \textbf{Solder feeder error.}
The solder feeder continuously reports its operating status through dedicated digital I/O signals. Errors such as feeder jams are detected directly from the corresponding error output.
\end{itemize}

We evaluated the framework's robustness through dedicated stress tests by manually inducing each anomaly at the corresponding stage of the automation cycle (\href{https://youtu.be/KW-GrarzYkI}{Supplementary Movie 7}). For each anomaly case, Supplementary Table~\ref{supp_tab:anomaly_cases} summarizes its occurrence rate during factory deployment, detection rate, mean recovery time, and mean post-production correction time. Detection rates were evaluated over 60 trials consisting of both nominal and anomalous conditions. Recovery and correction times were measured during factory deployment when the corresponding anomaly occurred naturally and through dedicated stress tests otherwise.

The framework detected most evaluated anomalies in stress tests. The measured recovery times suggest little additional production overhead. Periodic human inspection remains necessary, particularly for soldering failures (78\% detection rate) and unforeseen anomalies. Operators inspect completed motors during unloading every 12--15~min. These checks help limit the accumulation of production failures. A dedicated solder-joint classifier could replace the current insertion-based heuristic to improve soldering failure detection.

\subsection{Development and Design Process}\label{supp_sec:development_trials}

Our design process was guided by three critical principles applied across both the system- and component-level designs:
\begin{enumerate}
    \item \textbf{Hybrid control through task decomposition:} Decompose the workflow into subtasks, retain deterministic control for structured and repeatable subtasks, and apply learned control only where adaptation is necessary.
    \item \textbf{Minimizing data collection through task-specific inductive biases:} Structure task formulations, action spaces, and observations to reduce the amount of training data required.
    \item \textbf{Reducing avoidable downtime:} Incorporate monitoring and automatic stop-and-go mechanisms to maintain reliable operation while minimizing unnecessary interruptions.
\end{enumerate}

At the system level, these principles motivated a hybrid, task-decomposed architecture rather than a single end-to-end policy. At the component level, they guided the iterative selection and refinement of the approaches summarized in Supplementary Table~\ref{supp_tab:development_alternatives}, including both explored alternatives and adopted solutions.

\subsection{Additional Implementations on Different Hardware and Task}\label{supp_sec:additional_impl}
\subsubsection{Different Hardware Setup}
As a qualitative implementation example on a different hardware setup, we reproduced the same soldering automation process on a different workstation using a different robot platform (\href{https://youtu.be/DJctgNjhxyc}{Supplementary Movie 8}). The task scheduler, learning-based controller structure, and setup procedure were retained, while fixed waypoints were adjusted to account for the different hardware configuration. The limited task-specific data requirements of the learned components facilitated efficient migration of the learning-integrated system to the new hardware setup.

\subsubsection{Different Task: Chicken Sauce Brushing}
As a qualitative implementation example on a different task, we applied the learning-augmented automation framework to a chicken sauce-brushing task (\href{https://youtu.be/Te-zRx3jHxs}{Supplementary Movie 9}).

In this task, chicken pieces were randomly placed on a tray, with substantial variation in size, shape, and pose across instances. A dual-arm robot was required to grasp each piece individually in a suitable pose for processing, apply sauce to one surface, reorient the piece, and then brush the opposite side. The task was difficult to automate using conventional rule-based programming because of large part-to-part variation and difficulty of handcrafting brushing motions that adapt to different piece geometries.

We implemented the task using the same learning-augmented automation design. Visual servoing was used for chicken grasping, imitation learning for adaptive brushing motions, and conventional teaching-based control for structured subtasks including piece reorientation, piece release, and sauce loading onto the brush. As in the soldering system, learning was introduced selectively where perception-driven adaptation was required, while deterministic control was retained for repeatable motion segments.

This implementation is intended to illustrate how the same hybrid design can be applied for a task with different objects and manipulation requirements, rather than serving as a rigorous validation of task generality.

\begin{table}[H]
\centering
\small
\caption{Hyperparameters for the learning-based task controllers}
\label{supp_tab:task_controller_hparams}
\begin{tabular}{|ll|cc|}
\hline
\multicolumn{2}{|l|}{}                                                            & \multicolumn{1}{c|}{Visual Servoing} & Imitation Learning \\ \hline
\multicolumn{1}{|l|}{\multirow{12}{*}{Train}} & batch size                        & \multicolumn{2}{c|}{128}                                  \\ \cline{2-4} 
\multicolumn{1}{|l|}{}                        & learning rate                     & \multicolumn{2}{c|}{$1e-5$}           \\ \cline{2-4} 
\multicolumn{1}{|l|}{}                        & weight decay                      & \multicolumn{2}{c|}{$1e-4$}           \\ \cline{2-4} 
\multicolumn{1}{|l|}{}                        & vision backbone                   & \multicolumn{2}{c|}{Resnet18}                             \\ \cline{2-4} 
\multicolumn{1}{|l|}{}                        & num encoder layers                & \multicolumn{2}{c|}{4}                                    \\ \cline{2-4} 
\multicolumn{1}{|l|}{}                        & num decoder layers                & \multicolumn{2}{c|}{1}                                    \\ \cline{2-4} 
\multicolumn{1}{|l|}{}                        & num heads                         & \multicolumn{2}{c|}{8}                                    \\ \cline{2-4} 
\multicolumn{1}{|l|}{}                        & hidden dim                        & \multicolumn{2}{c|}{512}                                  \\ \cline{2-4} 
\multicolumn{1}{|l|}{}                        & feedforward dim                   & \multicolumn{2}{c|}{3200}                                 \\ \cline{2-4} 
\multicolumn{1}{|l|}{}                        & orientation representation & \multicolumn{2}{c|}{6D~\cite{zhou2019rotation}}                                   \\ \cline{2-4} 
\multicolumn{1}{|l|}{}                        & chunk size                        & \multicolumn{1}{c|}{1}               & 10                 \\ \cline{2-4} 
\multicolumn{1}{|l|}{}                        & CVAE usage                        & \multicolumn{1}{c|}{False}           & True               \\ \hline
\multicolumn{1}{|l|}{\multirow{2}{*}{Deploy}} & chunk size                        & \multicolumn{1}{c|}{1}               & 5                  \\ \cline{2-4} 
\multicolumn{1}{|l|}{}                        & success threshold                 & \multicolumn{1}{c|}{-}               & 0.95               \\ \hline
\end{tabular}
\end{table}

\begin{table}[H]
\centering
\small
\caption{Hyperparameters for the learning-based 3D safety monitor}
\label{supp_tab:safety_monitor_hparams}
\begin{tabular}{|l|l|l|c|}
\hline
\multirow{14}{*}{Train} & \multirow{7}{*}{Sim}                      & num environments           & 130                                                      \\ \cline{3-4} 
                        &                                           & episode length {[}s{]}     & 8                                                        \\ \cline{3-4} 
                        &                                           & box size {[}m{]}           & $U(0.3, 0.5) \times U(0.03, 0.35) \times U(0.03, 0.35)$  \\ \cline{3-4} 
                        &                                           & box position {[}m{]}       & $U(-0.65, 0.45) \times U(-1.45, 1.1) \times U(0., 1.85)$ \\ \cline{3-4} 
                        &                                           & LiDAR noise {[}m{]}        & $U(0, 0.005)$                                            \\ \cline{3-4} 
                        &                                           & LiDAR drift {[}m{]}        & $U(-0.01, 0.01)$                                         \\ \cline{3-4} 
                        &                                           & LiDAR update delay {[}s{]} & $U(0., 0.075)$                                           \\ \cline{2-4} 
                        & \multirow{2}{*}{Real}                     & data length                & 7 min 23 sec                                             \\ \cline{3-4} 
                        &                                           & co-train ratio             & 1:1                                                      \\ \cline{2-4} 
                        & \multirow{3}{*}{\makecell[l]{Model \\ architecture}} 
                                                                    & type                       & PointNet Encoder + Tri-plane Decoder                     \\ \cline{3-4} 
                        &                                           & exteroception hidden dim   & 8                                                        \\ \cline{3-4} 
                        &                                           & proprioception hidden dim  & 128                                                      \\ \cline{2-4} 
                        & \multirow{2}{*}{\makecell[l]{Area \\ definition}} 
                                                                    & grid region {[}m{]}        & $[-0.55, 0.35] \times [-1.35, 1.] \times [0.1, 1.75]$    \\ \cline{3-4} 
                        &                                           & grid resolution             & 0.05                                                     \\ \hline
\multirow{8}{*}{Deploy} & \makecell[l]{Stop zone \\ - fixed}   & grid region {[}m{]}        & $[-0.55, 0.35] \times [-1., 0.65] \times [0.4, 1.15]$    \\ \cline{2-4} 
                        & \multirow{7}{*}{\makecell[l]{Stop zone \\ - dynamic}} 
                        & $v_h$                       & 2        \\ \cline{3-4} 
                        &                                           & $t_r$                       & 0.1      \\ \cline{3-4} 
                        &                                           & $t_s$                       & 0.01     \\ \cline{3-4} 
                        &                                           & $b$                          & 0.005    \\ \cline{3-4} 
                        &                                           & $C$                          & 0.21     \\ \cline{3-4} 
                        &                                           & $z_r$                       & 0.001    \\ \cline{3-4} 
                        &                                           & $z_s$                       & 0.05     \\ \hline
\end{tabular}
\end{table}


\newpage

\end{document}